\documentclass[journal]{IEEEtran}

\ifCLASSINFOpdf
\else
\fi
\usepackage{graphicx}
\usepackage{amsmath}
\usepackage{amssymb}
\usepackage{multirow}
\usepackage{enumerate}
\usepackage{algorithm}
\usepackage{algorithmic}
\usepackage{slashbox}
\usepackage{array}
\usepackage{lineno,hyperref}
\usepackage{bm}
\usepackage{hyperref}
\usepackage{subfig}
\usepackage{epstopdf}

\begin{document}

\title{Exploiting Web Images for Dataset Construction:\\ A Domain Robust Approach}

\author{Yazhou~Yao,~\IEEEmembership{Student Member,~IEEE,}
	Jian~Zhang,~\IEEEmembership{Senior Member,~IEEE,}
	Fumin~Shen,~\IEEEmembership{Member,~IEEE,}
	Xiansheng~Hua,~\IEEEmembership{Fellow,~IEEE,}
	Jingsong~Xu,
	and~Zhenmin~Tang

\thanks{Y. Yao, J. Zhang and J. Xu are with the Global Big Data Technologies Center, University of Technology Sydney, NSW 2007, Australia.}
\thanks{F. Shen is with the School of Computer Science and Engineering, University of Electronic Science and Technology of China.}
\thanks{X. Hua is a researcher/senior director in Alibaba Group, Hangzhou, China.}
\thanks{Z. Tang is with the School of Computer Science and Engineering, Nanjing University of Science and Technology, China.}
\thanks{Corresponding author: Fumin Shen (Email: fumin.shen@gmail.com).}}
\maketitle

\begin{abstract}
Labelled image datasets have played a critical role in high-level image understanding. However, the process of manual labelling is both time-consuming and labor intensive. To reduce the cost of manual labelling, there has been increased research interest in automatically constructing image datasets by exploiting web images. Datasets constructed by existing methods tend to have a weak domain adaptation ability, which is known as the ``dataset bias problem''. To address this issue, we present a novel image dataset construction framework that can be generalized well to unseen target domains. Specifically, the given queries are first expanded by searching the Google Books Ngrams Corpus to obtain a rich semantic description, from which the visually non-salient and less relevant expansions are filtered out. By treating each selected expansion as a ``bag'' and the retrieved images as ``instances'', image selection can be formulated as a multi-instance learning problem with constrained positive bags. We propose to solve the employed problems by the cutting-plane and concave-convex procedure (CCCP) algorithm. By using this approach, images from different distributions can be kept while noisy images are filtered out. To verify the effectiveness of our proposed approach, we build an image dataset with 20 categories. Extensive experiments on image classification, cross-dataset generalization, diversity comparison and object detection demonstrate the domain robustness of our dataset.
\end{abstract}

\begin{IEEEkeywords}
Domain robust, multiple query expansions, image dataset construction, MIL
\end{IEEEkeywords}

\IEEEpeerreviewmaketitle

\section{Introduction}

\begin{figure}[tbp]
	\centering
	\includegraphics[width=0.43\textwidth]{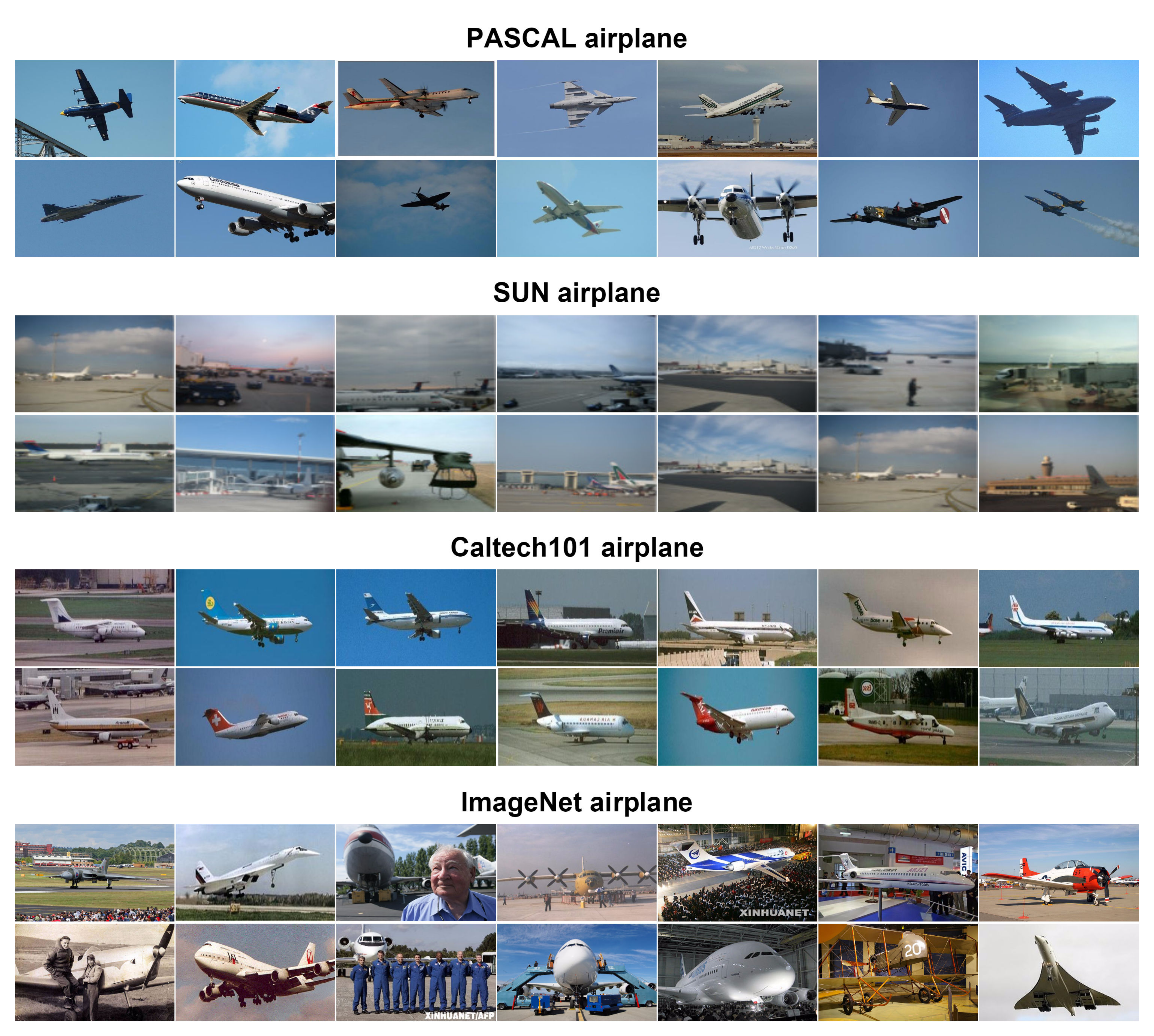}
	\caption{Most discriminative images for ``airplane" from four different datasets. Each dataset has their preference for image selection.}
	\label{fig1}  
\end{figure}

\IEEEPARstart{W}ith the development of the Internet, we have entered the era of big data. It is consequently a natural idea to leverage the large scale yet noisy data on the web for various vision tasks \cite{li2015weakly,zhang2014weakly,shen2013approximate,tang2016weakly,shen2016fast,shen2016face}. Methods of exploiting web images for automatic image dataset construction have recently become a hot topic \cite{hua2015prajna,schroff2011harvesting,icme2016yao,li2010optimol} in the field of multimedia processing. Existing methods \cite{hua2015prajna,schroff2011harvesting,li2010optimol} usually use an iterative mechanism in the process of image selection. However, due to the visual feature distribution of images selected in this way, these datasets tend to have the dataset bias problem \cite{niu2015visual,torralba2011unbiased,mm2016yao}. 
Fig. \ref{fig1} shows the ``airplane" images from four different image datasets. We can observe some significant differences in these datasets: PASCAL \cite{everingham2010pascal} shows ``airplanes" from the flying viewpoint, while SUN \cite{xiao2010sun} tends to show distant views at the airport; Caltech \cite{griffin2007caltech} has a strong preference for side views and 
ImageNet \cite{deng2009imagenet} is rich in diversity, but mainly contains close-range views. Classifiers learned from these datasets usually perform poorly in domain adaptation tasks \cite{niu2015visual}. 

To address the dataset bias problem, a large number of domain-robust approaches have been proposed for various vision tasks \cite{vijayanarasimhan2008keywords,duan2011improving,shen2015hashing}. The images in these methods are partitioned into a set of clusters; each cluster is treated as a ``bag" and the images in each bag as ``instances". As a result, these tasks can be formulated as a multi-instance learning (MIL) problem. Different MIL methods have been proposed in \cite{vijayanarasimhan2008keywords,duan2011improving}. However, the yield for all of these methods is limited by the diversity of training data which was collected with a single query.

To obtain highly accurate and diverse candidate images, as well as to overcome the download restrictions of the image search engine, Divvala et al. \cite{divvala2014learning} and Yao et al. \cite{icme2016yao,yao2016new} proposed the use of multiple query expansions instead of a single query to collect candidate images from the image search engine. The issue remains that these methods still use iterative mechanisms in the process of image selection, which leads to the dataset bias problem \cite{niu2015visual,torralba2011unbiased,mm2016yao}. 

Motivated by the situation described above, we target the construction of an image dataset in a scalable way, while ensuring accuracy and robustness. The basic idea is to leverage multiple query expansions for initial candidate images collection and to use MIL methods for selecting images from different distributions. We first expand each query to a set of query expansions, from which the visually non-salient and less relevant expansions are filtered out. Then we set the rest as selected query expansions and construct the raw image dataset with these selected query expansions. By treating each selected query expansion as a ``bag" and the images therein as ``instances", we formulate image selection and noise removal as a multi-instance learning problem. In this way, images from different distributions will be kept while noise is filtered out. 

To verify the effectiveness of our proposed approach, we build an image dataset with 20 categories, which we refer to as
DRID-20. We compare the image classification ability, cross-dataset generalization ability and diversity of our dataset with three manually labelled datasets and three automated datasets, to demonstrate the domain robustness of our dataset. We also report the results of object detection on PASCAL VOC 2007, and then compare the object detection ability of our method with weakly supervised and web-supervised methods.

The main contributions of this work are summarized as follows:

[1.] To the best of our knowledge, this is the first proposal for automatic domain-robust image dataset construction. Our proposed approach, based on multiple query expansions and multi-instance learning, considers the source of candidate images and retains images from different distributions. The dataset constructed by our approach thus efficiently alleviates the dataset bias problem. 

[2.] To suppress the search error and noisy query expansions induced noisy images, we formulate image selection as a multi-instance learning problem and propose to solve the associated optimization problems by the cutting-plane and concave-convex procedure (CCCP) algorithm, respectively.

[3.] We have released our image dataset DRID-20 on \href{https://drive.google.com/drive/folders/0B7dS7AFpUzt1bmpwcFRKcDZwUUE?usp=sharing}{Google Drive}. We hope the diversity of DRID-20 will offer unparalleled opportunities to researchers in the multi-instance learning, transfer learning, image dataset construction and other related fields.

This paper is an extended version of \cite{mm2016yao}. The extensions include: Taking both bag level and instance level noisy images into account in the process of image selection instead of only instance level noisy images, we use a combination of bag level and instance level selection mechanisms and achieve better results; comparing the image classification ability and diversity of our dataset DRID-20 with three manually labelled datasets (STL-10, CIFAR-10 and ImageNet) and three automated datasets (Optimol, Harvesting and AutoSet); and increasing the number of categories in the dataset from 10 to 20, so that our dataset DRID-20 covers all categories in the PASCAL VOC 2007 dataset.

The rest of the paper is organized as follows: In Section \uppercase\expandafter{\romannumeral2}, a brief discussion of related works is given. The proposed algorithm including query expanding, noisy expansions filtering and noisy images filtering is described in Section \uppercase\expandafter{\romannumeral3}. We evaluate the performance of the proposed algorithm against several other methods in Section \uppercase\expandafter{\romannumeral4}. Lastly, the conclusion and future work are offered in Section \uppercase\expandafter{\romannumeral5}.

\section{Related works}

Given the importance of labelled image datasets in the area of high-level image understanding, many efforts have been directed toward image dataset construction. In general, these efforts can be divided into three principal categories: manual annotation, active learning and automatic methods.  

\subsection{Manual Annotation and Active Learning methods}

In the early years, manual annotation was the most important way to construct image datasets. (e.g., STL-10 \cite{coates2011analysis}, CIFAR-10 \cite{krizhevsky2009learning}, PASCAL VOC 2007 \cite{everingham2010pascal}, ImageNet \cite{deng2009imagenet} and Caltech-101 \cite{griffin2007caltech}). The process of constructing these datasets mainly consists of submitting keywords to an image search engine to download candidate images, then cleaning these candidate images by manual annotation. This method usually has a high accuracy, but is labor intensive. 

To reduce the cost of manual annotation, a large number of works have focused on active learning (a special case of semi-supervised method). Li et al. \cite{collins2008towards} randomly labelled some seed images to learn visual classifiers. The learned visual classifiers were then implemented to conduct image classification on unlabelled images, to find low confidence images for manual labelling. Here low confidence images are those whose probability is classified into positive and negative close to 0.5. The process is iterated until sufficient classification accuracy is achieved. Siddiquie et al. \cite{siddiquie2010beyond} presented an active learning framework to simultaneously learn contextual models for scene understanding tasks (multi-class classification). Grauman et al. \cite{vijayanarasimhan2014large} presented an approach for on-line learning of object detectors, in which the system automatically refines its models by actively requesting crowd-sourced annotations on images crawled from the web. 

However, both manual annotation and active learning require pre-existing annotations, which often results in one of the most significant limitations to construct a large scale image dataset.

\subsection{Automatic Methods}

To further reduce the cost of manual annotation, automatic methods have attracted more and more people's attention. Schroff et al. \cite{schroff2011harvesting} adopted text information to rank images retrieved from a web search and used these top-ranked images to learn visual models to re-rank images once again. Li et al. \cite{li2010optimol} leveraged the first few images returned from an image search engine to train the image classifier, classifying images as positive or negative. When the image is classified as a positive sample, the classifier uses incremental learning to refine its model. With the increase in the number of positive images accepted by the classifier, the trained classifier will reach a robust level for this query. Hua et al. \cite{hua2015prajna} proposed to use clustering based method to filter ``group'' noisy images and propagation based method to filter individual noisy images. The advantage of these methods is that the need for manual intervention is eliminated. However, for methods \cite{schroff2011harvesting,li2010optimol,hua2015prajna}, the domain adaptation ability is limited by the initial candidate images and the iterative mechanism in the process of image selection. In order to obtain a variety of candidate images, Yao et al. \cite{icme2016yao} proposed the use of multiple query expansions instead of a single query in the process 
\begin{figure*}[htb]
	\centering
	\includegraphics[width=0.95\textwidth]{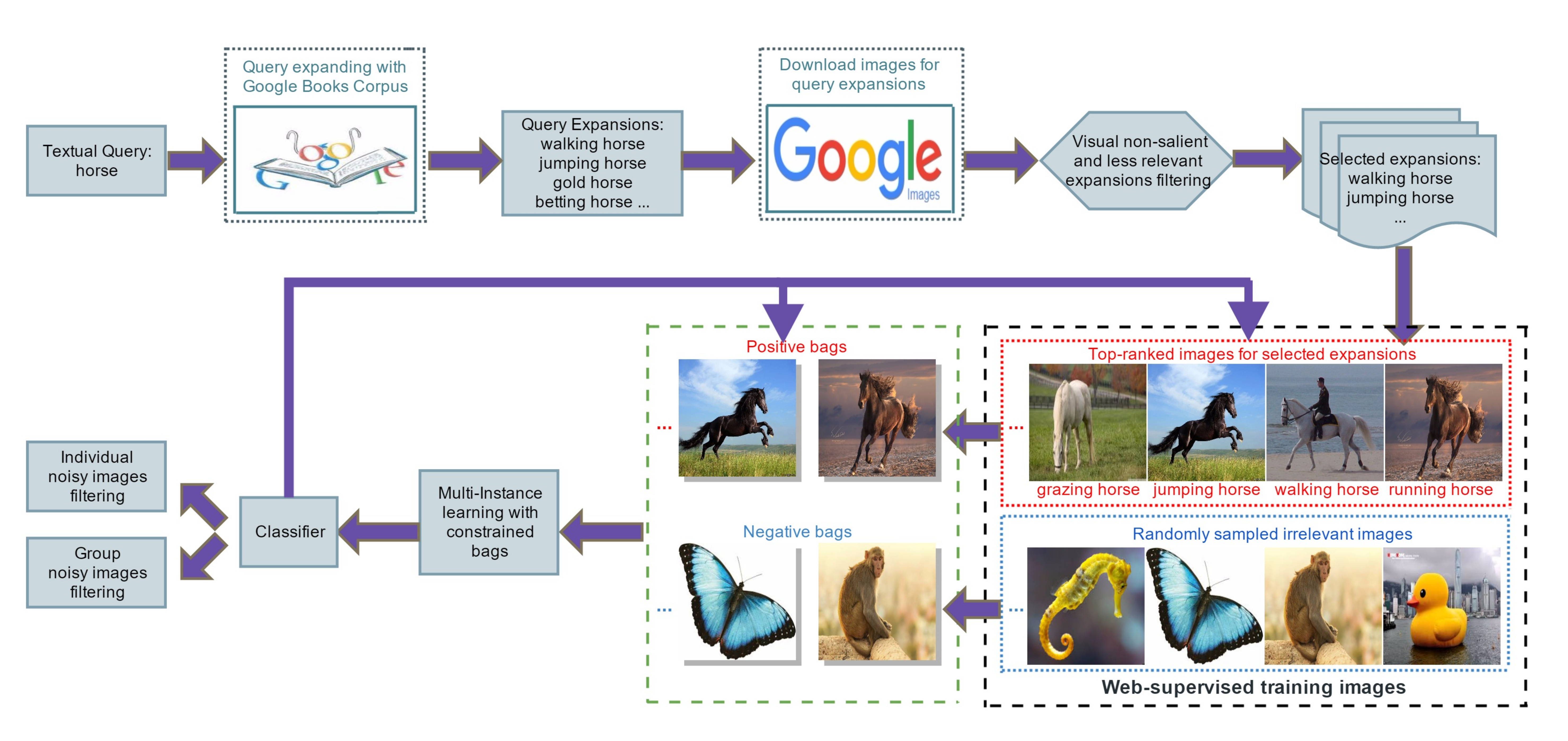}
	\caption{Domain robust image dataset construction framework. The input is text query that we would like to build a image dataset for. The outputs are a set of selected images corresponding to the given query.}
	\label{fig2}  
\end{figure*}
of initial candidate images collection, then using an iterative mechanism to filter noisy images. The automatic works discussed here mainly focus on accuracy and scale in the process of image dataset construction, which often results in a poor performance on domain adaptation.

\subsection{Other Related Works}

There are many works related to the generation of query expansions and noisy images filtering, though they are not aimed at image dataset construction. Since most image search engines restrict the number of images returned for each query, WordNet \cite{miller1995wordnet} and ConceptNet \cite{speer2013conceptnet} are often used to obtain synonyms to overcome the download restriction of these search engines. The advantage of WordNet and ConceptNet is that synonyms are usually relevant to the given query and almost do not need to be purified. The disadvantage of WordNet and ConceptNet is that both of them are usually not comprehensive enough for query expanding. Worse, the images returned from image search engine using synonyms tend to experience the homogenization problem, which results in poor performance on domain adaptation. 

In order to alleviate the homogenization problem, recent works \cite{icme2016yao,divvala2014learning} proposed the use of Google Books Ngram Corpus (GBNC) \cite{lin2012syntactic} instead of WordNet and ConceptNet to obtain query expansions for candidate images collection. The Google Books Ngrams Corpus covers almost all related queries at the text level. It is much more general and richer than WordNet and ConceptNet. The disadvantage of using GBNC for query expanding is that it may also bring noisy query expansions. Recently, word embedding \cite{cilibrasi2007google} provides a learning-based method for computing the word-word similarity distance which can be used to filter noisy query expansions. In this paper, we use GBNC to expand the query to a set of query expansions, and then take both word-word and visual-visual similarity to filter noisy query expansions.

To efficiently ease the dataset bias problem, several authors have developed domain-robust approaches for vision tasks. Duan et al. \cite{duan2011improving} clustered relevant images using both textual and visual features. By treating each cluster as a ``bag" and the images in the bag as ``instances", the authors formulated this problem as a multi-instance learning problem (MIL) which learns a target decision function for image re-ranking. However, the yield is limited by the initial candidate images obtained from image search engine with a single query. In this paper, we focus on the MIL method, as it can retain images from different distributions with noisy images filtered out.

It can be anticipated that there will be more visual patterns (corresponding to different query expansions) in our work to represent the given query. In addition, MIL methods are applied to filter noisy images and keep images from different distributions. In return, the constructed dataset could achieve a better domain adaptation ability than traditional datasets constructed by a single query and an iterative mechanism.

\section{Domain robust image dataset construction}

We seek to construct a domain-robust image dataset that can generalize to unseen target domains. As shown in Fig. \ref{fig2}, we propose our web-supervised image dataset construction framework by three major steps: query expanding, noisy query expansions filtering and noisy images filtering. We expand the query to a set of semantically rich expansions by searching Google Books Ngram Corpus, from which the visually non-salient and less relevant expansions are filtered out. After obtaining the candidate images by retrieving the selected expansions with an image search engine, we treat each selected expansion as a ``bag" and the images in each bag as ``instances". We then formulate image selection and noisy images filtration as an MIL problem with constrained positive bags. In particular, the learned classifiers are used to filter individual noisy images (corresponding to the top-ranked images for selected expansions) and group noisy images (corresponding to the positive bags). Using this approach, images from different distributions will be kept while noisy images are filtered out, and a domain-robust image dataset will be constructed.

\subsection{Query Expanding}

Image datasets constructed by existing methods tend to be highly accurate but usually have weak domain adaptation ability \cite{niu2015visual,torralba2011unbiased,mm2016yao}. To construct a domain-robust image dataset, we expand given query (e.g., ``horse'') to a set of query expansions (e.g., ``jumping horse, walking horse, roaring horse'') and then use these different query expansions (corresponding images) to reflect the different ``visual distributions" of the query. We use GBNC to discover query expansions for the given query with Parts-Of-Speech (POS), specifically with NOUN, VERB, ADJECTIVE and ADVERB. GBNC is much more general and richer than WordNet \cite{miller1995wordnet} and ConceptNet \cite{speer2013conceptnet}. Using GBNC can help us to find all the expansions ever published for any possible query from the text semantics \cite{divvala2014learning}.

\subsection{Noisy Expansions Filtering}

Through query expanding, we obtain a comprehensive semantic description for the given query. However, query expanding not only provides all the useful query expansions, but also some noise. These noisy query expansions can be roughly divided into two types: (1) visually non-salient (e.g., ``betting horse'') and (2) less relevant (e.g., ``sea horse''). Using these noisy query expansions to retrieve images will have a negative effect on dataset accuracy and robustness. 

\subsubsection{Visual non-salient expansions filtering}

From the visual perspective, we aim to identify visually salient and eliminate non-salient query expansions in this step. The intuition is that visually salient expansions should exhibit predictable visual patterns. Hence, we can use the image classifier-based filtering method.
For each query expansion, we directly download the top $N$ images from the Google image search engine as positive images ( based on the fact that the top few images returned from image search engine tend to be positive), then randomly split these images into a training set and validation set $I_i=\{I_i^t, I_i^v\}$. We gather a random pool of negative images and split them into a training set and validation set $\overline I=\{\overline I^t, \overline I^v\}$. We train a linear support vector machine (SVM) classifier $C_i$ with $I_i^t$ and $\overline I^t$ using dense histogram of oriented gradients (HOG) features \cite{dalal2005histograms}. We then use $\{I_i^v,\overline I^v\}$ as validation images to calculate the classification results. We declare a query expansion $i$ to be visually salient if the classification results $S_i$ give a relatively high score.

\subsubsection{Less relevant expansions filtering} 

From the relevance perspective, we want to identify both semantically and visually relevant expansions for the given query. The intuition is that relevant expansions should have a relatively small semantic and visual distance; therefore, we use a combined word-word and visual-visual similarity distance-based filtering method.
Words and phrases acquire meaning from the way they are used in society. For computers, the equivalent of ``society'' is ``database'', and the equivalent of ``use'' is ``a way to search the database'' \cite{cilibrasi2007google}. Normalized Google Distance (NGD) constructs a method to extract semantic similarity distance from the World Wide Web (WWW) using Google page counts. For a search term $x$ and search term $y$, NGD is defined by:
\begin{equation}\label{eq1}
\mathrm{NGD}(x,y)=\frac{\max\{\log f(x),\log f(y)\}-\log f(x,y)}{\log N-\min\{\log f(x),\log f(y)\}}
\end{equation}
where $f(x)$ denotes the number of pages containing $x$, $f(x,y)$ denotes the number of pages containing both $x$ and $y$ and $N$ is the total number of web pages searched by Google.

We denote the semantic distance of all query expansions by a graph $G_{semantic} = \{N, D\}$ in which each node represents a query expansion and its edge represents the NGD between two nodes. We set the target query as center $ y $ and other expansions have a score $D_{xy}$ which corresponds the NGD to the target query.
Similarly, we represent the visual distance of query and expansions by a graph $G_{visual} = \{C,E\}$ in which each node represents a query expansion and each edge represents the visual distance between the query and the expansions. We denote the visual distribution of each query expansion by the compound feature $\phi_{k}=\frac{1}{k}\sum_{i=1}^{k}x_{i}$ of its first $k$ images from the image search engine. We set the target query as center $ y $ and other query expansions have a score $E_{xy}$ which corresponds to the Euclidean distance to the target query. 

The semantic distance $D_{xy}$ and visual distance $E_{xy}$ will be used to construct a new two-dimensional feature  $V = [D_{xy};E_{xy}]$. The problem is to calculate the importance weight $w$ and bias penalty $b$ in decision function $f(x) = w^{T}x+b$ to determine whether or not the expansion is relevant. There are many methods of obtaining these coefficients $w$ and $b$. Here we take the linear SVM to work around this problem. Although the linear SVM is not the prevailing state-of-the-art method for classification, we find our method to be effective in pruning irrelevant query expansions.

We set the remainder which is not filtered out as the selected expansions and construct raw image dataset by retrieving the top $N$ images from image search engine with these selected query expansions.
Regardless of the fact that our method is not able to remove noisy expansions thoroughly in most cases, the raw image dataset constructed by our method still achieves much higher accuracy than directly using the Google image data. Besides, the raw image dataset constructed through the selected query expansions has much richer visual distributions.

\subsection{Noisy Images Filtering}

Although the Google image search engine has ranked the returned images, some noisy images may still be included. In addition, a few noisy expansions which are not filtered out will also bring noisy images to the raw image dataset. In general, these noisy images can be divided into two types: group noisy images (caused by noisy query expansions) and individual noisy images (as a result of the error index of the image search engine). To filter these group and individual noisy images while retaining the images from different distributions, we use MIL methods instead of iterative methods in the process of image selection and noise removal.

By treating each selected expansion as a ``bag" and the images corresponding to the expansion as ``instances", we formulate a multi-instance learning problem by selecting a subset of bags and a subset of images from each bag to construct the domain-robust image dataset. Since the precision of images returned from image search engine tends to be relatively high, we define each positive bag as at least having a portion of $\delta $ positive instances which effectively filter group noisy images caused by noisy query expansions.

We denote each instance as $x_{i}$ with its label $y_{i} \in \left \{ 0,1 \right \}$, where $ i $ =1,...,$ n $. We also denote the label of each bag $B_{I}$ as $Y_{I} \in \left \{ 0,1 \right \}$. The transpose of a vector or matrix is represented by superscript $'$ and the element-wise product between two matrices is represented by $\odot$. We define the identity matrix as $\mathbf{I}$ and $\mathbf{0}$, $\mathbf{1}$ $\in \Re ^{n}$ denote the column vectors of all zeros and ones, respectively. The inequality $\mathbf{u}=\left [ u_{1},u_{2}...u_{n} \right ]{}'\geq \mathbf{0}$ means that $u_{i}\geq 0$ for $ i $ =1,...,$ n $.

\subsubsection{Filtering individual noisy images}

The decision function for filtering individual noisy images is assumed in the form of $ f(x)={w}' \varphi (x)+b$ and has to be learned from the raw image dataset. We employ the formulation of Lagrangian SVM, in which the square bias penalty $b^{2}$ and the square hinge loss for each instance are used in the objective function. The decision function can be learned by minimizing the following structural risk function:
\begin{equation}\label{eq2}
\min_{y,w,b,\rho ,\varepsilon_{i}} \:
\frac{1}{2}\left ( \left \| w \right \|^{2}+b^{2}+C\sum_{i=1}^{n}\varepsilon _{i}^{2} \right )-\rho \\
\end{equation}
\begin{equation}\label{eq3}
\text{s.t.} \:\: y_{i}({w}'\varphi \left ( x_{i} \right )+b)\geq \rho -\varepsilon_{i}, i=1,...n. \\
\end{equation}
\begin{equation}\label{eq4}
\begin{aligned}
& & \sum_{i:x_{i}\in B_{I}} \frac{y_{i}+1}{2}\geq \delta \left | B_{I} \right | \quad for \quad Y_{I}=1,  \\
& & \quad y_{i}=0 \quad \quad for \quad Y_{I}=0 \\
\end{aligned}
\end{equation}
where $\varphi$ is a mapping function that maps $x$ from the original space into a high dimensional space $\varphi (x)$, $C>0$ is a regularization parameter and $\varepsilon _{i}$ values are slack variables. The margin separation is defined as $\rho /\left \| w \right \|$. $ y=\left [ y_{1}...y_{n} \right ]'$ means the vector of instance labels, $\lambda =\{ \mathbf{y} |y_{i}\in \left \{ 0,1 \right \} \}$ and $\mathbf{y}$ satisfies constraint (\ref{eq4}). By introducing a dual variable $\alpha _{i}$ for inequality constraint (\ref{eq3}) and kernel trick $k_{ij}=\varphi (x_{i}){}'\varphi (x_{j})$, we arrive at the optimization problem below:
\begin{equation}\label{eq5}
\min_{\mathbf{y}\in \lambda} \max_{\alpha} \:
-\frac{1}{2}(\sum_{i,j=1}^{n}\alpha_{i}\alpha_{j}y_{i}y_{j}k_{ij}+\sum_{i,j=1}^{n}\alpha_{i}\alpha_{j}y_{i}y_{j}+\frac{1}{C})  \\
\end{equation}
where $\alpha _{i}\geq 0$, $\sum_{i=1}^{n}\alpha _{i}=1$ and $\mathbf{\alpha }=\left [ \alpha_{1},\alpha_{2}...\alpha_{n} \right ]{}'$. By defining $\mathbf{K}=\left [ k_{ij} \right ]$ as a $n\times n$ kernel matrix, $\mathbf{\nu }=\left \{ \mathbf{\alpha }|\mathbf{\alpha \geq 0, \alpha {}'1}=1 \right \}$ and $\widetilde{\mathbf{K}}=\mathbf{K}+\mathbf{1}\mathbf{1}{}'$ as a $n\times n$ transformed kernel matrix for the augmented feature mapping $\widetilde{\varphi }(x)= [\varphi (x){}'. \: 1]{}'$ 
of kernel $\widetilde{k_{ij}}=\widetilde{\varphi}(x_{i}){}'\widetilde{\varphi}(x_{j})$. (\ref{eq5}) can be rewritten as follows:
\begin{equation}\label{eq6}
\min_{\mathbf{y}\in \lambda}\max_{\alpha \in \nu} \:
-\frac{1}{2}\alpha {}'(\widetilde{\mathbf{K}}\odot \mathbf{y}\mathbf{y}{}'+\frac{1}{C}\mathbf{I})\alpha. \\
\end{equation}

\begin{algorithm}[thb]
	\caption{Cutting-plane algorithm for solving (\ref{eq10})}
	\label{alg1}
	\begin{algorithmic}[1]
		\STATE Initialize $y_{i}=Y_{I}$ for $x_{i}\in B_{I}$ as $\mathbf{y}^1$, and set $ \zeta =\left \{ \mathbf{y}^1 \right \}$; \\
		\STATE Use MKL to solve $\alpha $ and $\mathbf{u}$ in (\ref{eq10}) with $\zeta$; \\	
		\STATE Select most violated $\mathbf{y}^{t}$ with $\alpha$ and set $\zeta =\mathbf{y}^{t}\cup \zeta$;\\
		\STATE Repeat step 2 and step 3 until convergence. \\			
	\end{algorithmic}
\end{algorithm}

(\ref{eq6}) is a mixed integer programming problem with respect to the instance labels $y_{i}\in \left \{ 0,1 \right \}$. We take the Label-Generating MMC (LG-MMC) algorithm proposed in \cite{li2009tighter} to solve this mixed integer programming problem.
We first consider interchanging the order of $\max_{\alpha \in \upsilon }$ and $\min_{\mathbf{y}\in \lambda}$ in (\ref{eq6}) and obtain:
\begin{equation}\label{eq7}
\max_{\alpha \in \nu}\min_{\mathbf{y}\in \lambda} \:
-\frac{1}{2}\alpha'(\widetilde{\mathbf{K}}\odot \mathbf{y}\mathbf{y}{}'+\frac{1}{C}\mathbf{I})\alpha. \\
\end{equation}
According to the minmax theorem \cite{kim2008minimax}, the optimal objective of (\ref{eq6}) is an upper bound of (\ref{eq7}). We rewrite (\ref{eq7}) as:
\begin{equation}\label{eq8}
\max_{\alpha \in \nu}\left \{\max_{\theta}-\theta |\theta \geq \frac{1}{2} \alpha '(\widetilde{\mathbf{K}}\odot \mathbf{y}^{t}\mathbf{y}^{t^{'}}+\frac{1}{C}\mathbf{I})\alpha , \forall \mathbf{y}^{t}\in \lambda \right  \}
\end{equation}
$\mathbf{y}^{t}$ is any feasible solution in $\lambda$. For the inner optimization sub-problem, let $u_{t}\geq 0$ be the dual variable for inequality constraint. Its Lagrangian can be obtained as:
\begin{equation}\label{eq9}
-\theta +\sum_{t:\mathbf{y}_{t}\in \lambda }u_{t}(\theta - \frac{1}{2} \alpha {}'(\widetilde{\mathbf{K}}\odot \mathbf{y}^{t}\mathbf{y}^{t^{'}}+\frac{1}{C}\mathbf{I})\alpha ).
\end{equation}
Setting the derivative of (\ref{eq9}) with respect to $\theta$ to zero, we have $\sum u_{t}=1$. $\mathbf{M}=\left \{ \mathbf{u} | \sum u_{t}=1,u_{t}\geq 0 \right \}$ is denoted as the domain of $\mathbf{u}$, where $\mathbf{u}$ is the vector of $u_{t}$. The inner optimization sub-problem is replaced by its dual and (\ref{eq8}) can be rewritten as:
\begin{equation*}
\max_{\alpha \in \nu}\min_{\mathbf{u}\in \mathbf{M}} \:
-\frac{1}{2}\alpha {}'(\sum _{t:\mathbf{y}^{t}\in \lambda }u_{t} \widetilde{\mathbf{K}}\odot \mathbf{y}\mathbf{y}{}'+\frac{1}{C}\mathbf{I})\alpha \\
\end{equation*}
or
\begin{equation}\label{eq10}
\min_{\mathbf{u}\in \mathbf{M}}\max_{\alpha \in \nu} \:
-\frac{1}{2}\alpha {}'(\sum _{t:\mathbf{y}^{t}\in \lambda }u_{t} \widetilde{\mathbf{K}}\odot \mathbf{y}\mathbf{y}{}'+\frac{1}{C}\mathbf{I})\alpha. \\
\end{equation}
Here, we can interchange the order of $\max_{\alpha \in \nu}$ and $\min_{\mathbf{u}\in \mathbf{M}}$ because the objective function is concave in $\mathbf{\alpha }$ and convex in $\mathbf{u}$. Additionally, (\ref{eq10}) can be regarded as a multiple kernel learning (MKL) problem \cite{bach2004multiple}, and the target kernel matrix is a convex combination of base kernel matrices $\left \{ \widetilde{\mathbf{K}}\odot \mathbf{y_{t}}\mathbf{y_{t}}{}' \right \}$.
Although $\lambda$ is finite and (\ref{eq10}) is an MKL problem, we can not directly use existing MKL techniques like \cite{rakotomamonjy2008simplemkl} to solve this problem. The reason is that the exponential number of possible labellings $\mathbf{y}_{t}\in \lambda$ and the fact that the base kernels are exponential in size make direct MKL computations intractable.

Fortunately, not all the constraints in (\ref{eq8}) are active at optimality, thus we can employ a cutting-plane algorithm \cite{kelley1960cutting} to find a subset $\zeta \in \lambda$ of the constraints that can well approximate the original optimization problem. The detailed solutions of the cutting-plane algorithm for (\ref{eq10}) are described in Algorithm \ref{alg1}. Finding the most violated constraint $\mathbf{y}^{t}$ is the most challenging aspect of the cutting-plane algorithm.

According to (\ref{eq5}), the most violated $\mathbf{y}^{t}$ is equivalent to the following optimization problem:
\begin{equation}\label{eq11}
\max_{\mathbf{y}\in \lambda} \:
\sum_{i,j=1}^{n}\alpha _{i}\alpha _{j}y_{i}y_{j}k_{ij}. \\
\end{equation}
We solve this integer optimization problem by enumerating all possible candidates of $\mathbf{y}^{t}$. Here we only enumerate the possible labelling candidates of the instances in positive bags as all instances in the negative bags are assumed to be negative in our paper. Lastly, we can derive the decision function from the raw image dataset for the given query as:
\begin{equation}\label{eq12}
f(x)=\sum _{i:\alpha _{i}\neq 0}\alpha _{i}\widetilde{y}_{i}\widetilde{k}(x,x_{i}) \\
\end{equation}
where $\widetilde{y}_{i}= \sum _{t:\mathbf{y}^{t}\in \lambda }u_{t}y_{i}^{t}$ and $\widetilde{k}(x,x_{i})={k}(x,x_{i})+1$. The decision function will be used to filter individual noisy images in each bag which correspond to selected query expansions. 

\subsubsection{Filtering group noisy images}

To filter group noisy images, we represent bag $B_{I}$ with the compound feature $\phi_{f,k}$ of its first $k$ positive instances:
\begin{equation}\label{eq13}
\phi_{f,k}(B_{I})=\frac{1}{k}\sum_{x_{i}\in \Psi_{f,k}^{*}(B_{I})}x_{i}  \\
\end{equation}
with
\begin{equation}\label{eq14}
\Psi_{f,k}^{*}(B_{I})=_{\Psi\subseteq B_{I},|\Psi|=k}^{\quad \arg \max}\sum_{x_{i}\in \Psi}f(x_{i}). \\
\end{equation}
We refer to the instances in $\Psi_{f,k}^{*}(B_{I})$ as the first $k$ instances of $B_{I}$ according to classifier $f$ (see Equation \ref{eq12}). Since the closer of images in $B_{I}$ from the bag center, the higher probability of these images to be relevant to the bag. The assignment of relatively heavier weights to images which have short distance to bag center would increase the accuracy of classifying bag $B_{I}$ to be positive or negative, then increase the efficiency of filtering noisy group images. Following \cite{carneiro2007supervised}, we assume $\xi_{i}=[1+\exp(\alpha \log d(x_{i})+\beta )]^{-1}$ to be a weighting function, $d(x_{i})$ represents the Euclidean distance of images $x_{i}$ from the bag center, $\alpha \in \mathbb{R}_{++}$ and $\beta$ are scaling and offset parameters which can be determined by cross-validation. The representation of (\ref{eq13}) for bag $B_{I}$ can be generalized to a weighted compound feature:
\begin{equation}\label{eq15}
\phi_{f,k}(B_{I})=\phi(X,h^{*})=\frac{Xh^{*}}{\xi^{T}h^{*}} \\
\end{equation}
with
\begin{equation}\label{eq16}
h^{*}= _{\:\:\: h \in H}^{\arg \max}f(\frac{Xh}{\xi^{T}h}), \:\:\: \text{s.t.} \:\: \sum _{i}h_{i}= k\\
\end{equation}
where $X=\left [ x_{1},x_{2},x_{3}..,x_{i} \right ]\in \mathbb{R}^{D\times i}$ is a matrix whose columns are the instances of bag $B_{I}$, $\xi = \left [ \xi_{1},\xi_{2},\xi_{3}...\xi_{i} \right ]^{T}\in \mathbb{R}_{++}^{i}$ are the vectors of weights, and $h^{*}\in H=\left \{ 0,1 \right \}^{i}\setminus \left \{ 0 \right \}\left ( \ \sum _{i}h_{i}= k \right )$ is an indicator function for the first k positive instances of bag $B_{I}$.

Then classifying rule of bag $B_{I}$ to be selected or not is:
\begin{equation}\label{eq17}
f_{\omega}(X)=_{h\in H}^{\max}\omega^{T}\phi(X,h), \:\:\: \sum_{i}h_{i}=k\\
\end{equation}
where $\omega \in \mathbb{R}^{D}$ is the vector of classifying coefficients, $\phi(X,h)\in \mathbb{R}^{D}$ is the feature vector of (\ref{eq15}), $h$ is a vector of 
\begin{algorithm}[thb]
	\caption{Concave-convex procedure for solving (\ref{eq21})}
	\label{alg2}
	\begin{algorithmic}[1]
		\STATE Initialize $\omega$ with SVM by setting $h=\mathbf{1}\in \mathbb{R}^{i}$; \\
		\STATE Compute a convex upper bound using the current model for the second term of (\ref{eq21}); \\	
		\STATE Minimize this upper bound by solving a structural SVM problem via the proximal bundle method \cite{kiwiel1990proximity};\\
		\STATE Repeat step 2 and step 3 until convergence. \\			
	\end{algorithmic}
\end{algorithm}
latent variables and $H$ is the hypothesis space $\left \{ 0,1 \right \}^{i}\setminus \left \{ 0 \right \}$. 
The learning problem is to determine the parameter vector $\omega$.

Given a training set $\tau = \left \{ B_{I},Y_{I} \right \}_{I=1}^{n}$, this is a latent SVM learning problem:
\begin{equation}\label{eq18}
\min_{\mathbf{\omega }}\frac{1}{2}\left \| \omega  \right \|^2+C\sum_{I=1}^{n}\max\left ( 0,1-Y_{I}f_\omega \left ( X_{B_I} \right ) \right ).\
\end{equation} 
Before solving (\ref{eq18}), we first solve the classifying rule of (\ref{eq17}). It is necessary to solve the below following problem:
\begin{equation}\label{eq19}
_{h \in H}^{\max}\textrm{}\frac{\omega ^{T}Xh}{\xi^{T}h}, \:\:\: \text{s.t.} \:\: \sum_{i}h_{i}=k.\ 
\end{equation}
This is an integer linear-fractional programming problem. Since $\xi \in \mathbb{R}_{++}^{i}$, (\ref{eq19}) is identical to the relaxed problem:
\begin{equation}\label{eq20}
_{h\in \ss ^i}^{\max}\frac{\omega ^{T}Xh}{\xi^{T}h}, \:\:\: \text{s.t.} \:\: \sum_{i}h_{i}=k\
\end{equation} 
where $\ss ^i=\left [ 0,1 \right ]^i$ is a unit box in $\mathbb{R}^i$. (\ref{eq20}) is a linear-fractional programming problem and can be reduced to a linear programming problem of $i+1$ variables and $i+2$ constraints \cite{boyd2004convex}.

In this work, we take the concave-convex procedure (CCCP) \cite{yuille2003concave} algorithm to solve (\ref{eq18}). We rewrite the objective of (\ref{eq18}) as two convex functions:
\begin{equation}\label{eq21}
\begin{aligned}
& & \min_{\mathbf{\omega }}
\left [ \frac{1}{2}\left \| \omega  \right \|^2+C\sum_{I\in D_{n}}\max\left ( 0,1+f_\omega \left ( X_{B_I} \right ) \right ) +  \right. \\
& & \left. C\sum_{I\in D_{p}}\max\left ( f_\omega \left ( X_{B_I} \right ),1 \right ) \right ]-\left [ C\sum_{I\in D_{p}} f_\omega \left ( X_{B_I} \right ) \right ] \\
\end{aligned}
\end{equation}

where $D_{p}$ and $D_{n}$ are positive and negative training sets respectively. The detailed solutions of the CCCP algorithm for (\ref{eq21}) are described in Algorithm \ref{alg2}. Lastly, we obtain the bag classifying rule as (\ref{eq17}) to filter group noisy images which correspond to noisy query expansions.  

In summary, the existing automatic methods reduce the cost of manual annotation by leveraging the generalization ability of machine learning models. However, this generalization ability is affected by both the quality of the initial candidate images and the capability of models to retain images from different distributions. Previous works primarily focus on
accuracy and scale, and most use an iterative mechanism for the image selection process which often results in a dataset bias problem. To the best of our knowledge, this is the first proposal for automatic domain-robust image dataset construction. We achieve the domain adaptation ability of our dataset by maximizing both the initial candidate images and the final selected images from different distributions. 

\section{Experiments}

To demonstrate the effectiveness of our approach, we have constructed an image dataset with 20 categories. We compare the image classification ability, cross-dataset generalization ability and diversity of our dataset with three manually labelled and three automated datasets. The motivation is to verify that a domain-robust image dataset has a better image classification ability on a third-party dataset; and to confirm that a domain-robust image dataset has better cross-dataset generalization ability and dataset diversity. We also report the object detection ability of our dataset and compare it with weakly supervised and web-supervised state-of-the-art methods.

\subsection{Image Dataset DRID-20 Construction} 

Since most existing weakly supervised and web-supervised learning methods were evaluated on the PASCAL VOC 2007 dataset, we choose the 20 categories in PASCAL VOC 2007 as the target categories for the construction of DRID-20.

For each given query (e.g.,``horse"), we first expand the given query to a set of query expansions with POS. To filter
visual non-salient expansions, we retrieve the top $\bold{N} = 100$ images from the image search engine as positive images (in spite of the fact that noisy images might be included). Set the training set and validation set $I_i=\{I_i^t=75, I_i^v=25\}$, $\overline I=\{\overline I^t=25, \overline I^v=25\}$. By experimentation, we declare a query expansion $i$ to be visually salient if the classification result ($S_i \geq 0.7$) returns a relatively high score. We have released the query expansions for 20 categories in DRID-20 and the corresponding images (original image URL) on: \href{https://drive.google.com/drive/folders/0B7dS7AFpUzt1bmpwcFRKcDZwUUE?usp=sharing}{Google Drive}.

To filter the less relevant expansions, we select $n_{+}$ positive training samples from these expansions that have a small semantic or visual distance. We calculate the semantic distance and visual distance between the different queries (e.g., ``horse'' and ``cow'') to obtain the $n_{-}$ negative training samples. Here, we set $n=1000$ and train a classifier based on linear SVM to filter less relevant expansions. 

The first $\bold{N}=100$ (for category ``plant'' expansions, $\bold{N}=350$) images are retrieved from image search engine for each selected query expansion to construct the raw image dataset. We treat the selected query expansions as positive bags and images therein as instances. Specifically, we define each positive bag as having at least a portion of $\delta =0.7 $ positive instances. Negative bags can be obtained by randomly sampling a few irrelevant images. MIL methods are applied to learn the decision function (\ref{eq12}) for individual noisy images filtering. 
The decision function (\ref{eq12}) is also used to select the most $k$ positive instances in each bag, representing this bag for group noisy images filtering. The value of $k$ for different categories may be different. In general, categories with larger query expansions tend to select a smaller value. There are multiple methods for learning the weighting function (e.g., logistic regression or cross-validation), here we follow \cite{carneiro2007supervised} and use cross-validation to learn the weighting function. To this end, we label 10 datasets, each containing 100 positive bags and 100 negative bags. The positive bags and negative bags each have 50 images. Labelling only needs to be carried out once to learn the weighting function and the weighted bag classification rule (\ref{eq17}). The learned weighted bag classification rule (\ref{eq17}) will be used to filter noisy bags (corresponding to group noisy images). For better comparison with other datasets, we evenly select positive images from positive bags to construct the dataset DRID-20. Each category in DRID-20 has 1000 images and this dataset has been released publicly on \href{https://drive.google.com/drive/folders/0B7dS7AFpUzt1bmpwcFRKcDZwUUE?usp=sharing}{Google Drive}. 

\subsection{Comparison of Image Classification Ability, Cross-dataset Generalization Ability and Dataset Diversity}

The goal of these experiments is to evaluate the domain robustness of our dataset.

\subsubsection{Experimental setting}

\begin{figure*} [thb]
	\centering
	\subfloat[]{%
		\includegraphics[width=1.7in, height=1.6in]{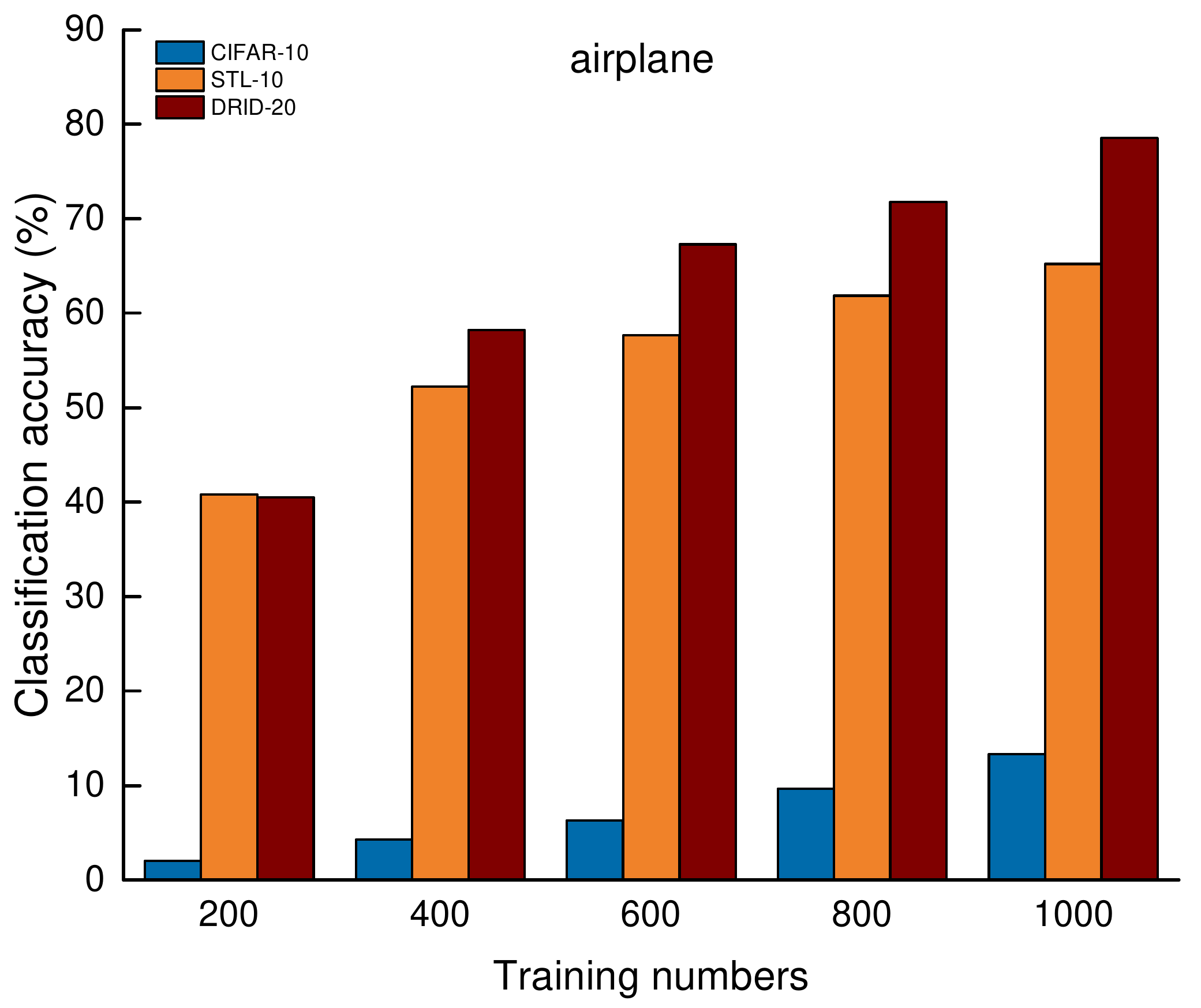}}
	\subfloat[]{%
		\includegraphics[width=1.7in, height=1.6in]{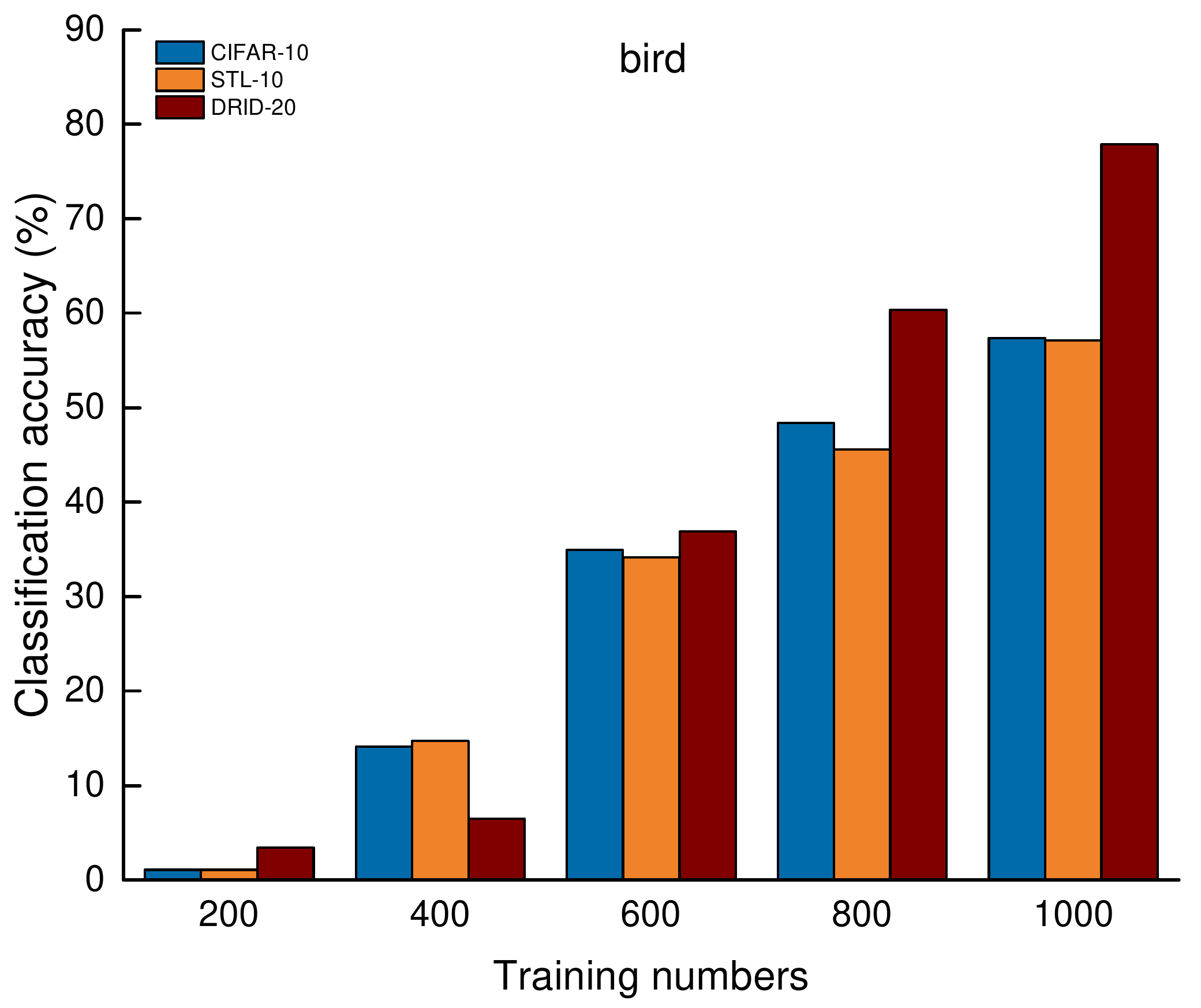}}
	\subfloat[]{%
		\includegraphics[width=1.7in, height=1.6in]{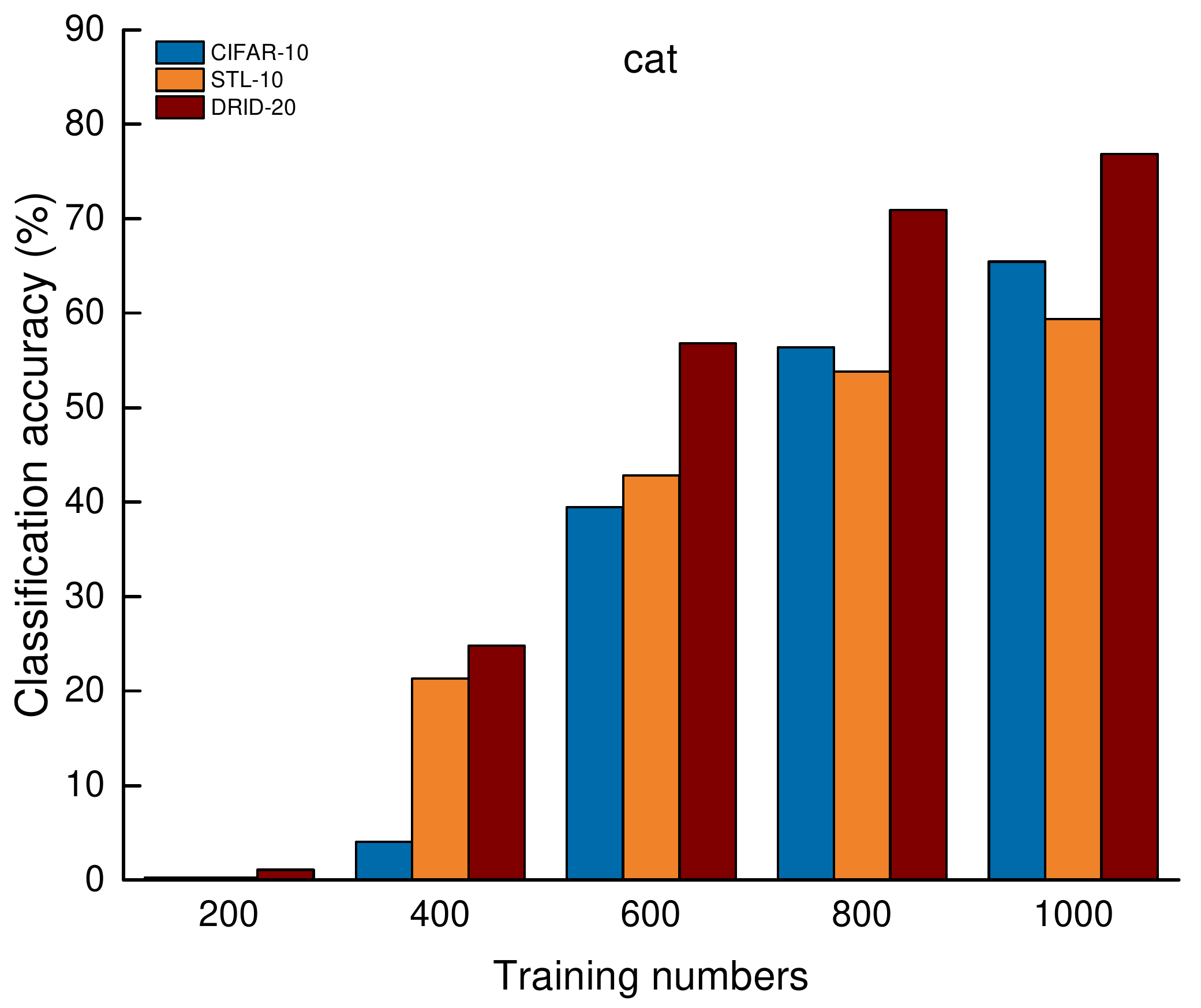}}
	\subfloat[]{%
		\includegraphics[width=1.7in, height=1.6in]{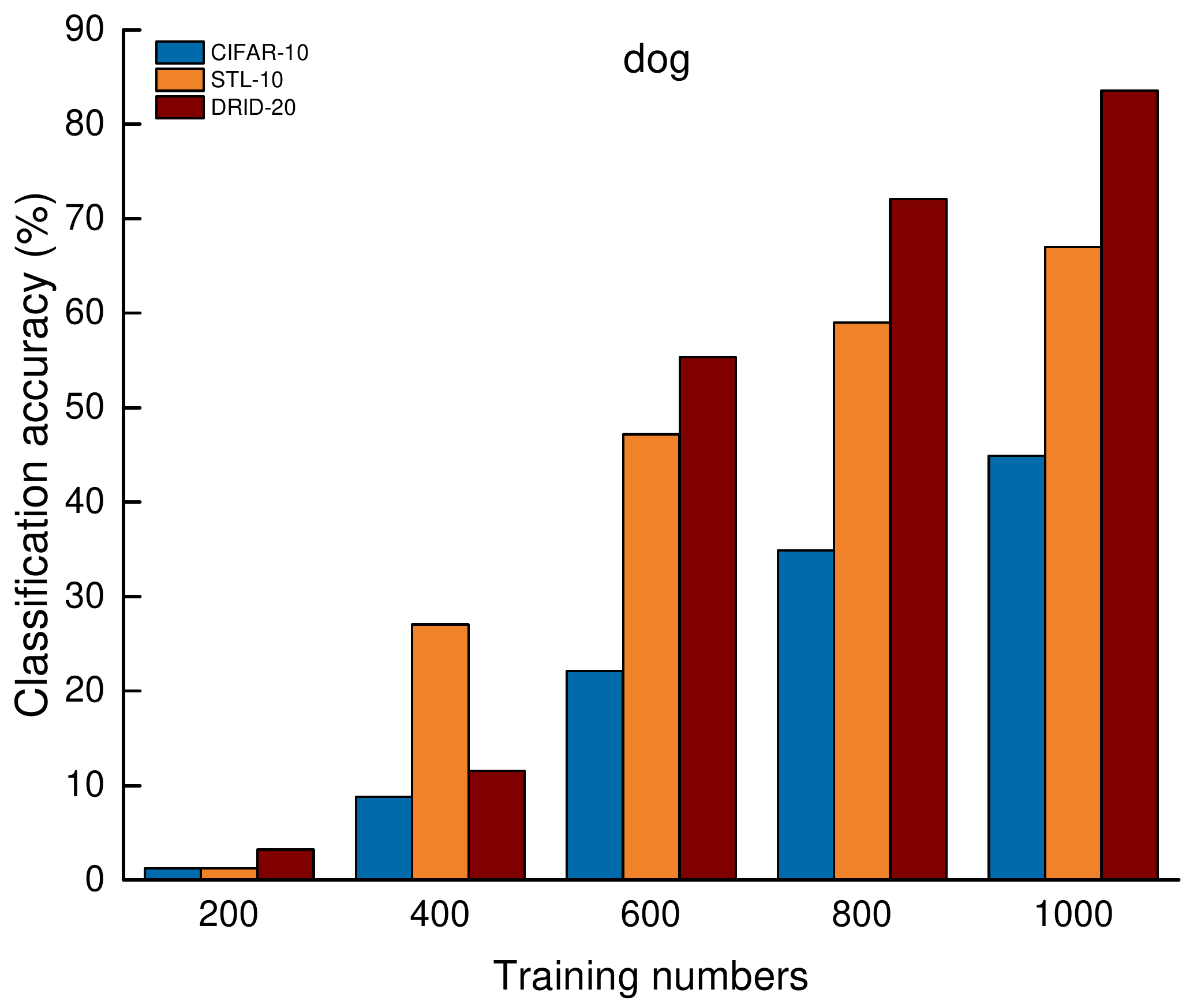}}\
	\subfloat[]{%
		\includegraphics[width=1.7in, height=1.6in]{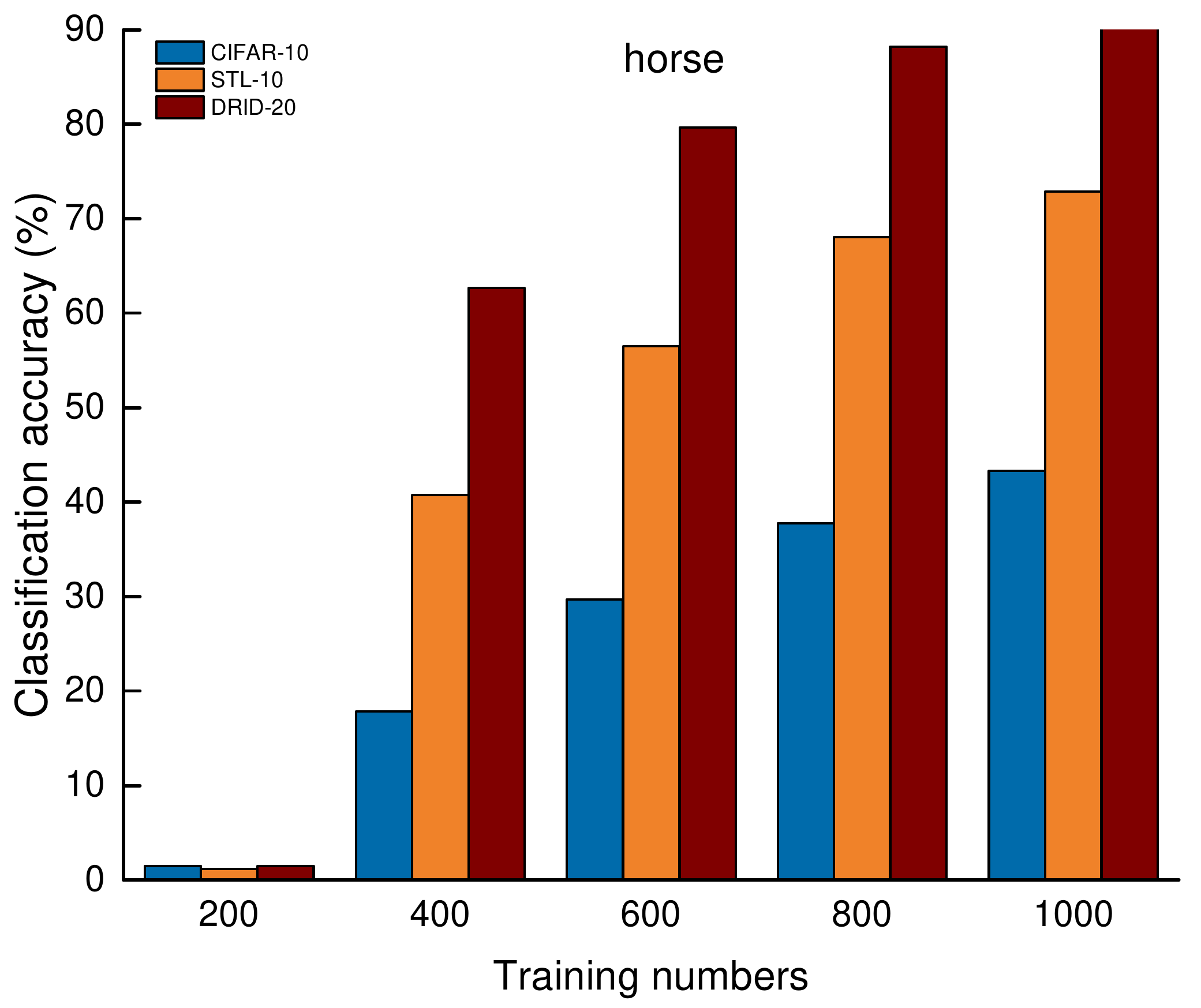}}
	\hspace{0.9cm}
	\subfloat[]{%
		\includegraphics[width=1.7in, height=1.6in]{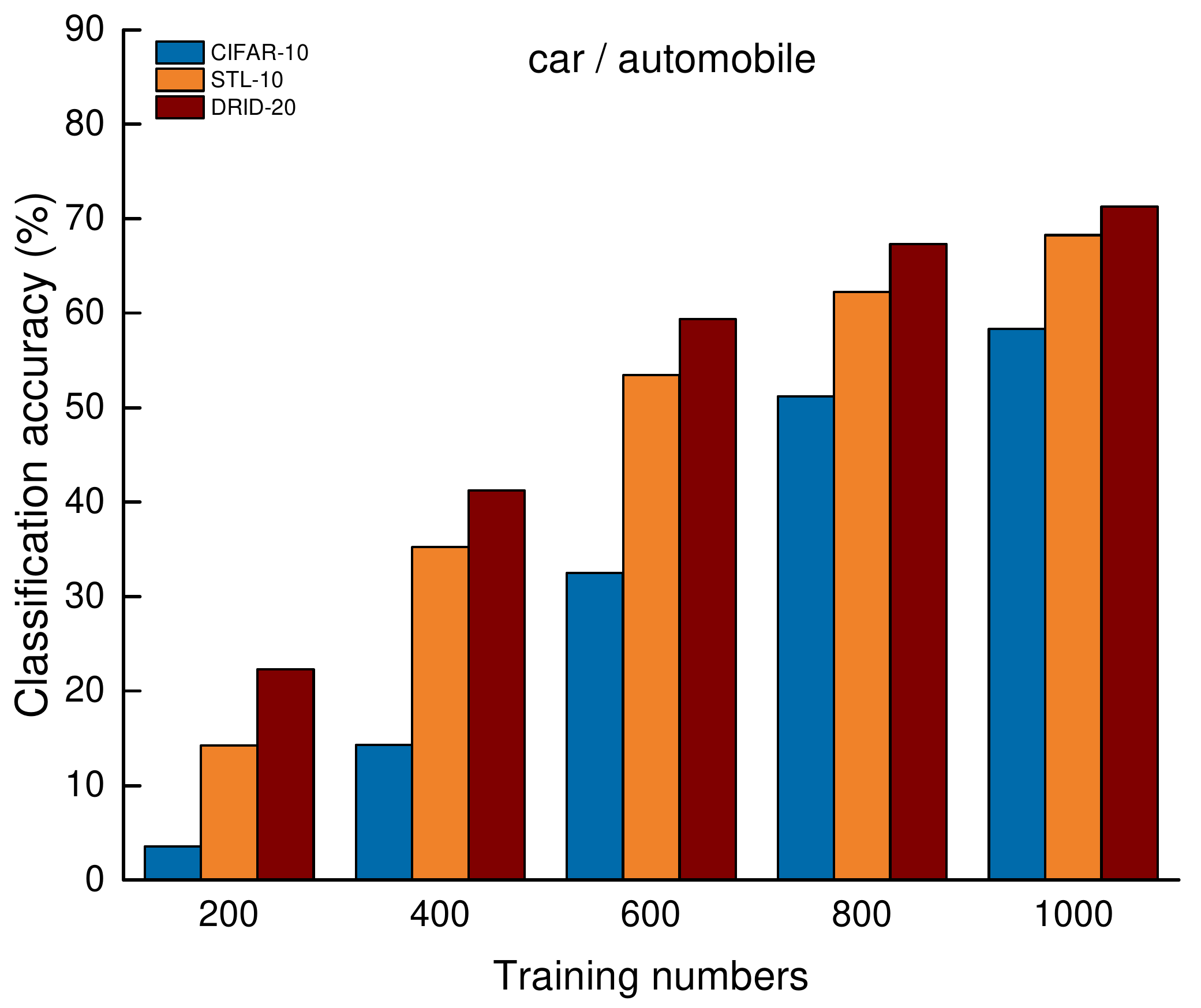}}
	\hspace{0.9cm}	
	\subfloat[]{%
		\includegraphics[width=1.7in, height=1.6in]{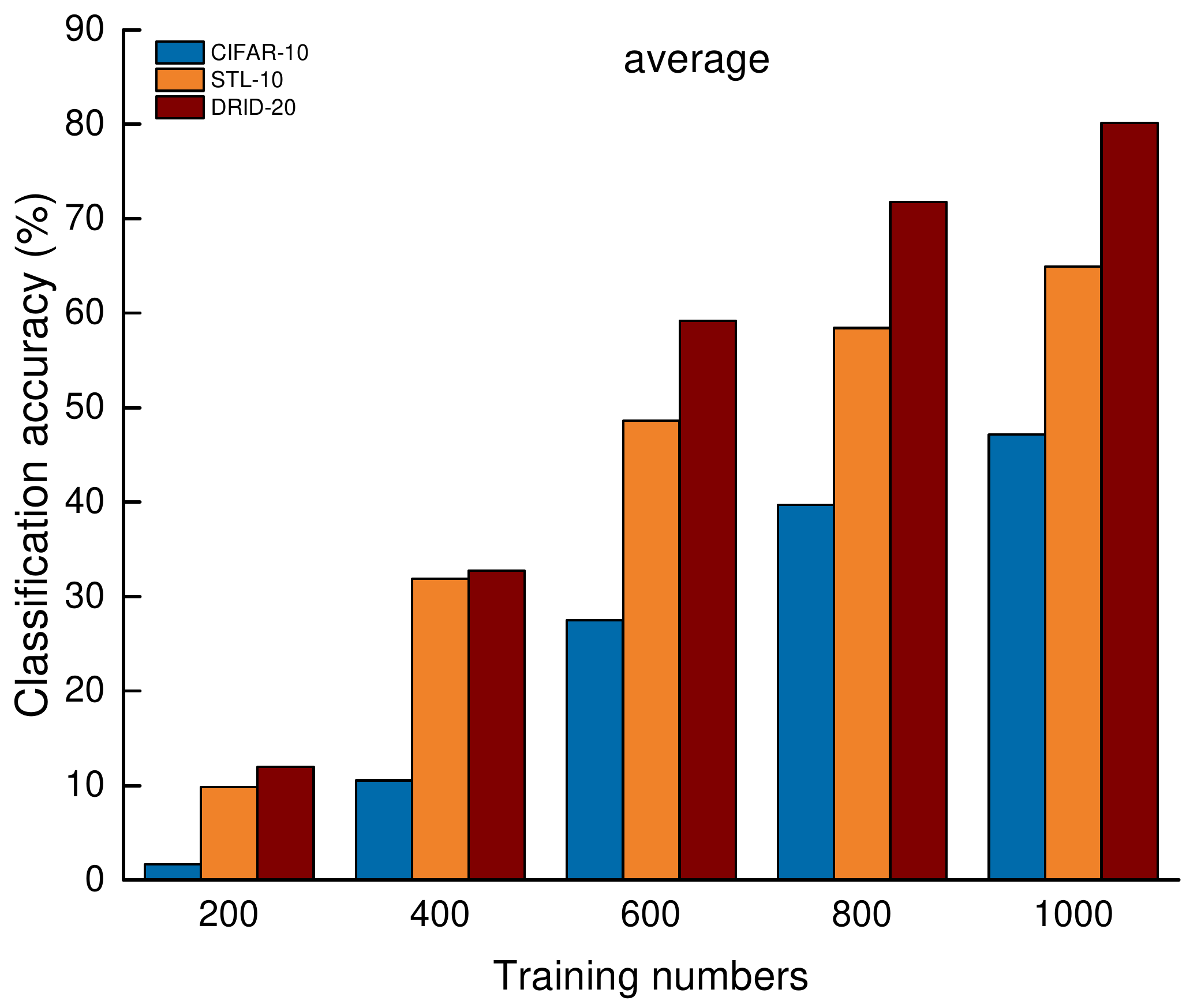}}
	\caption{Image classification ability of CIFAR-10, STL-10 and DRID-20 on PASCAL VOC 2007 dataset: (a) airplane, (b) bird, (c) cat, (d) dog, (e) horse, (f) car/automobile and (g) average.}
	\label{fig3}
\end{figure*}

\begin{figure*}[thb]
	\centering
	\includegraphics[width=7 in, height=2.08 in]{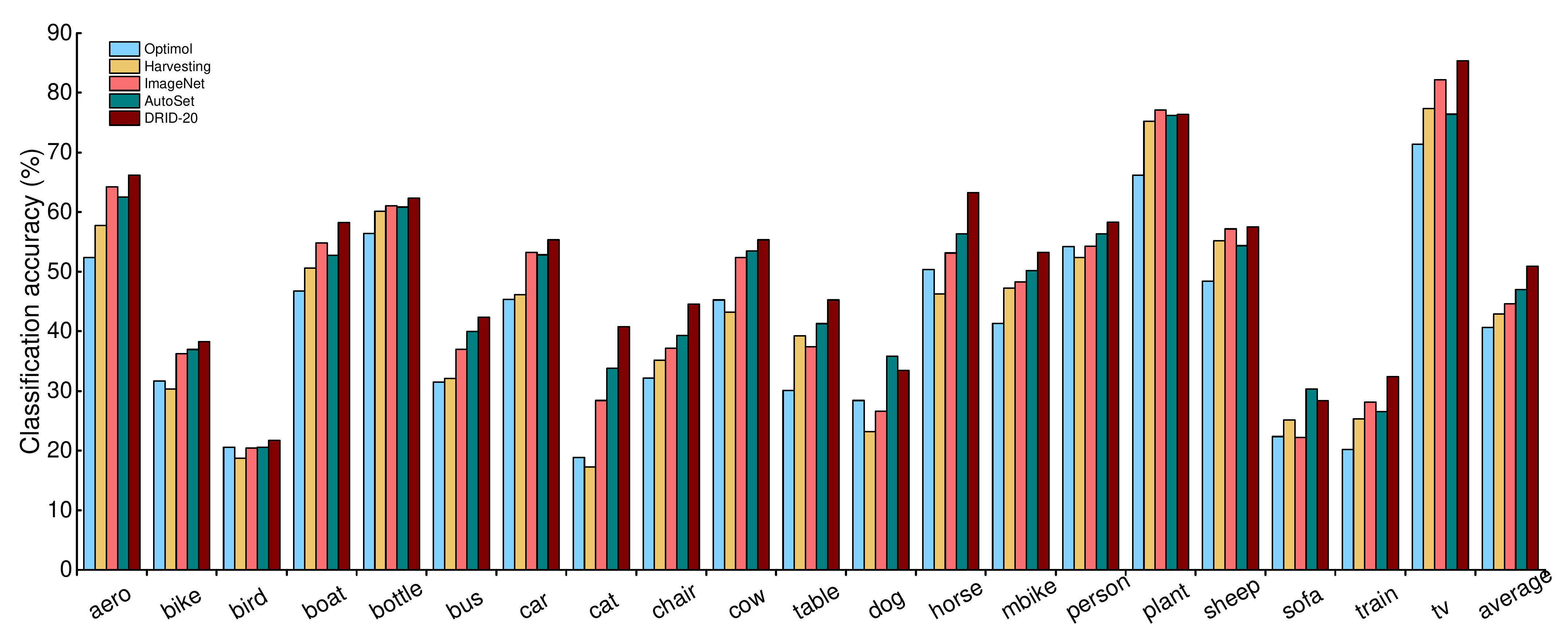}
	\caption{Image classification ability of Optimol, Harvesting, ImageNet, AutoSet and DRID-20 on PASCAL VOC 2007 dataset.}
	\label{fig4}  
\end{figure*}

We chose PASCAL VOC 2007 as the third-party testing benchmark dataset for comparing the image classification ability of our dataset with other baseline datasets. For this experiment, the same categories between various datasets are compared. Specifically, we compare the category ``airplane", ``bird", ``cat", ``dog", ``horse" and ``car/automobile" between STL-10\cite{coates2011analysis}, CIFAR-10 \cite{krizhevsky2009learning} and DRID-20. We sequentially select [200,400,600,800,1000] training images from CIFAR-10, STL-10 and DRID-20 as the positive training images, and use 1000 fixed irrelevant images as the negative training images to learn the image classifiers. 
For comparison with ImageNet \cite{deng2009imagenet}, Optimol \cite{li2010optimol}, Harvesting \cite{schroff2011harvesting} and AutoSet \cite{icme2016yao}, we use all the 20 categories among these datasets. In specific, we randomly select 500 training images for each category from these datasets as the positive training images. Similarly, we use 1000 fixed irrelevant images as the negative training images to learn the image classifiers. We then test the performance of these classifiers on the corresponding categories of the PASCAL VOC 2007 dataset. We repeat the above experiment ten times and use the average performance as the final performance for each dataset. The image classification ability of all datasets for each category is shown in Fig. \ref{fig3} and Fig. \ref{fig4}.

\begin{figure*}[t]
	\centering
	\subfloat[]{%
		\includegraphics[width=1.7in, height=1.6in]{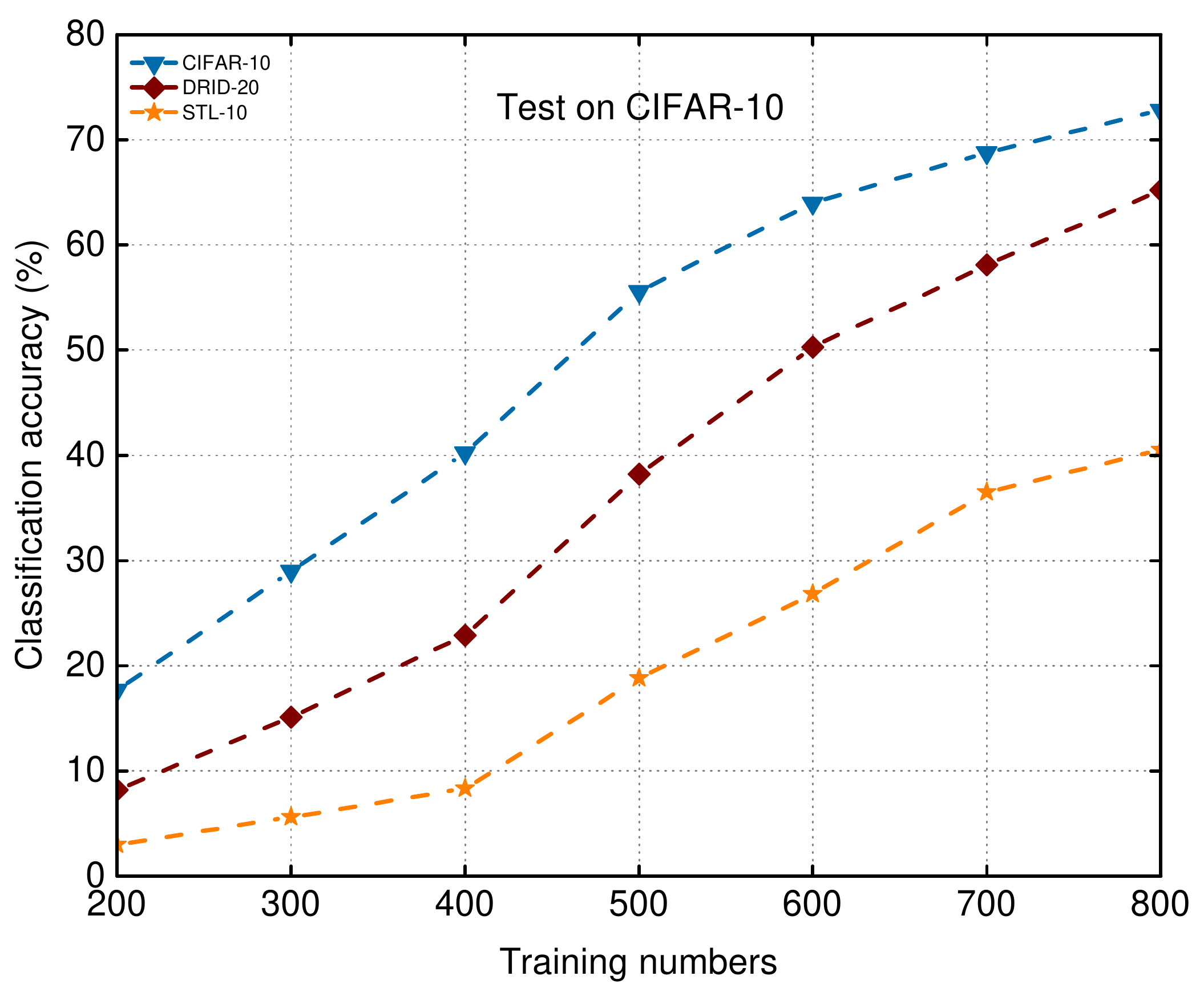}}
	\subfloat[]{%
		\includegraphics[width=1.7in, height=1.6in]{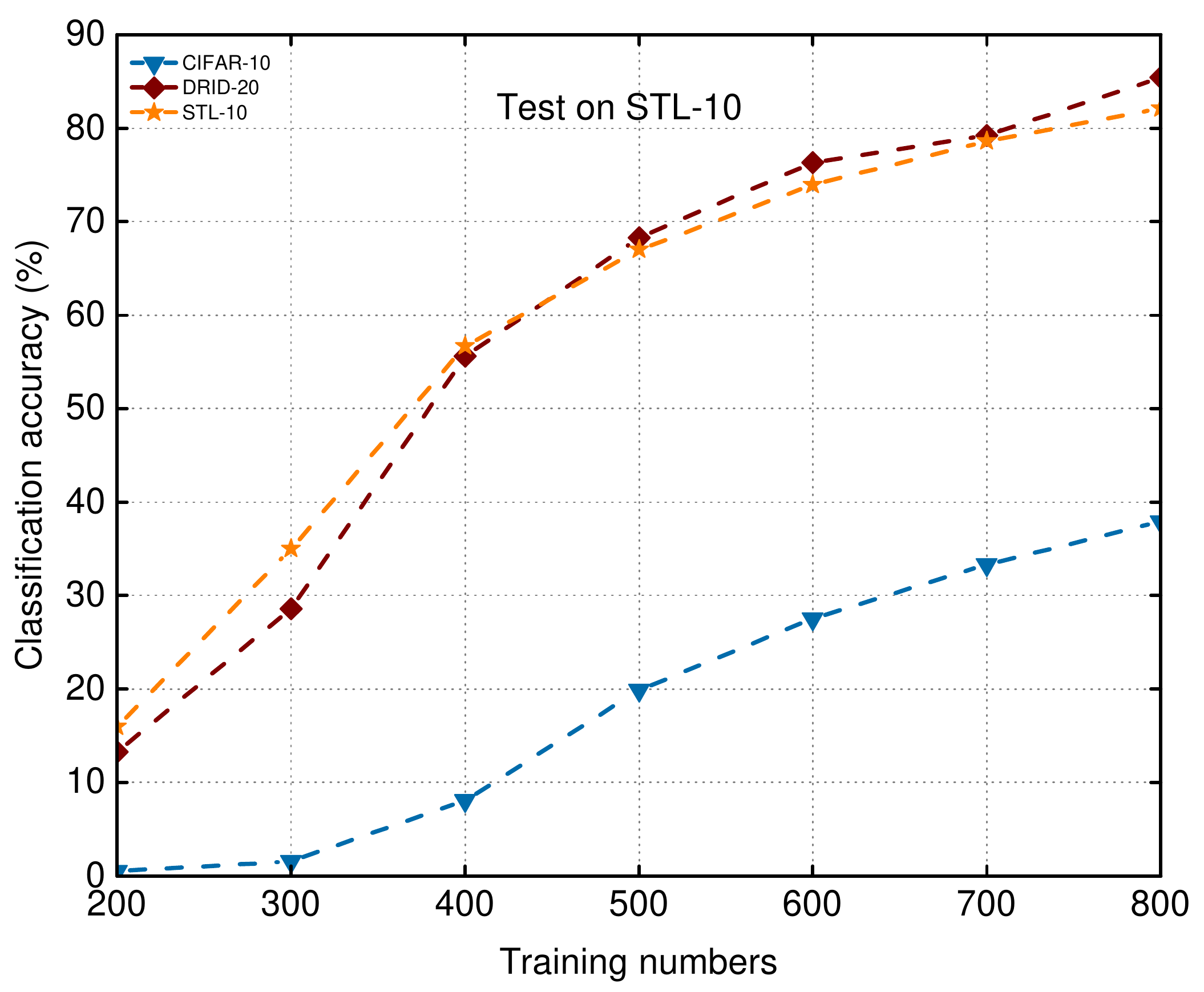}}\
	\subfloat[]{%
		\includegraphics[width=1.7in, height=1.6in]{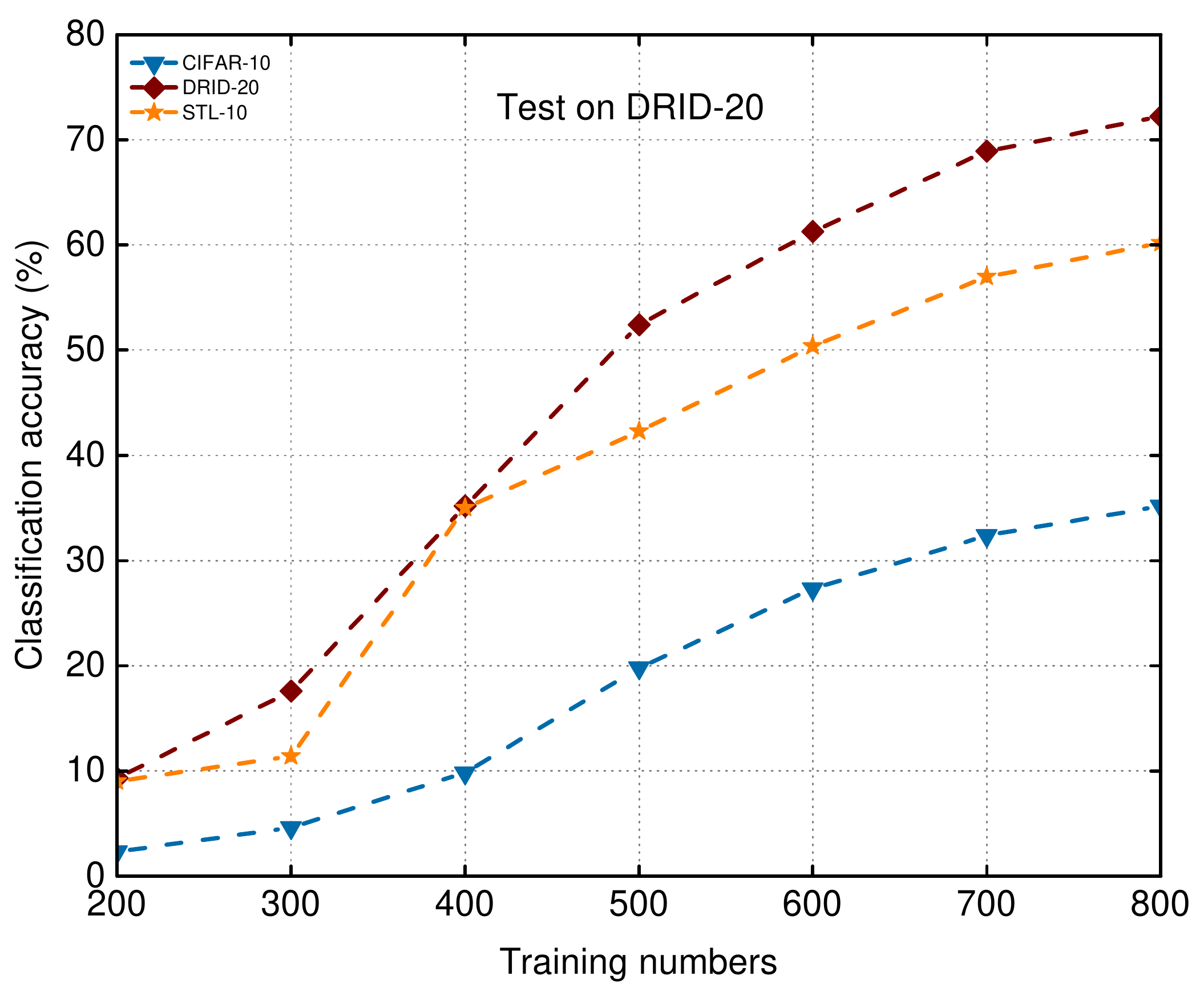}}
	\subfloat[]{%
		\includegraphics[width=1.7in, height=1.6in]{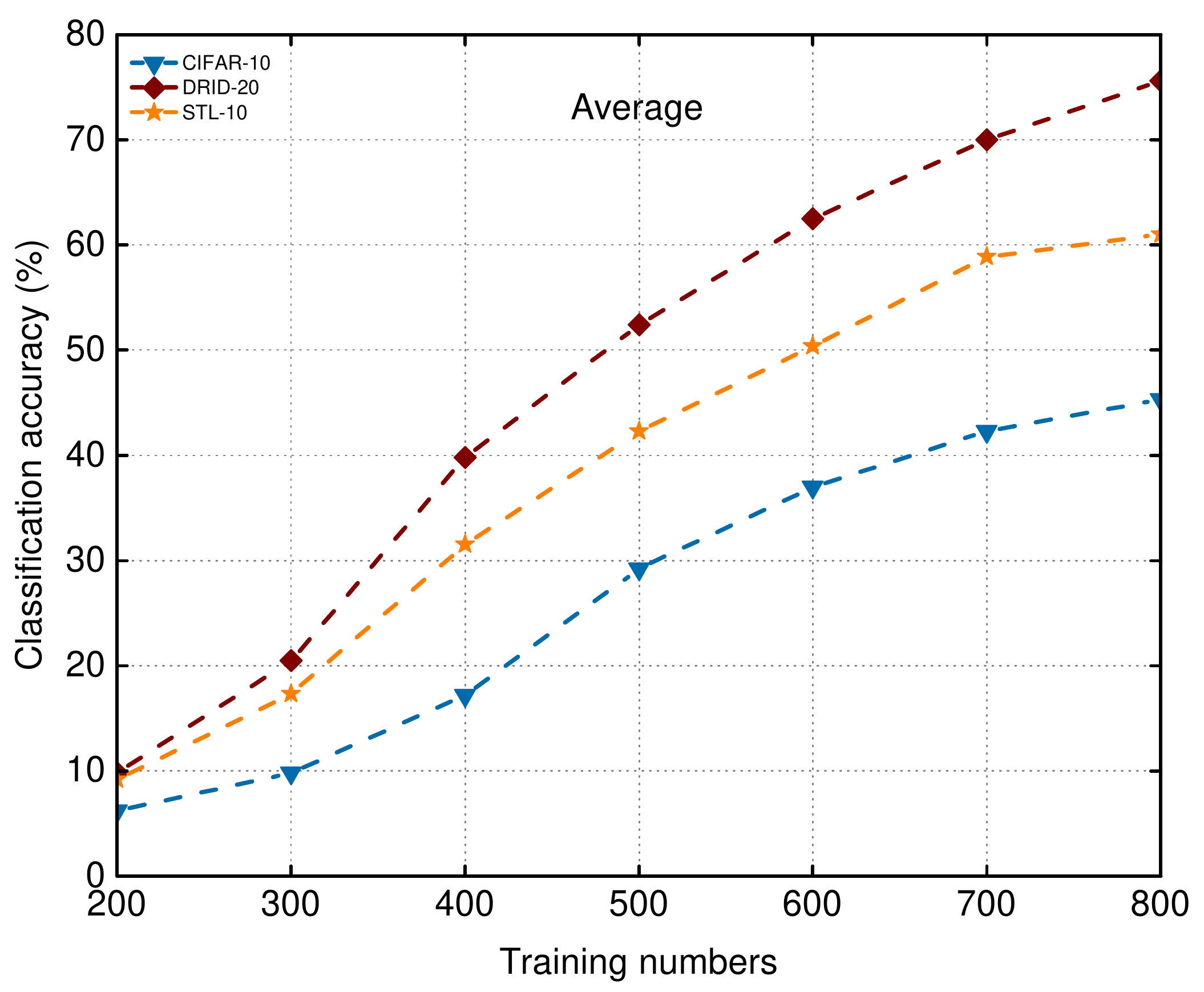}}
	\caption{Cross-dataset generalization ability of classifiers learned from CIFAR-10, STL-10, DRID-20 and then tested on: (a) CIFAR-10, (b) STL-10, (c) DRID-20, (d) Average.}
	\label{fig5}
\end{figure*}

\begin{figure*}[t] 
	\centering
	\subfloat[]{%
		\includegraphics[width=1.7in, height=1.6in]{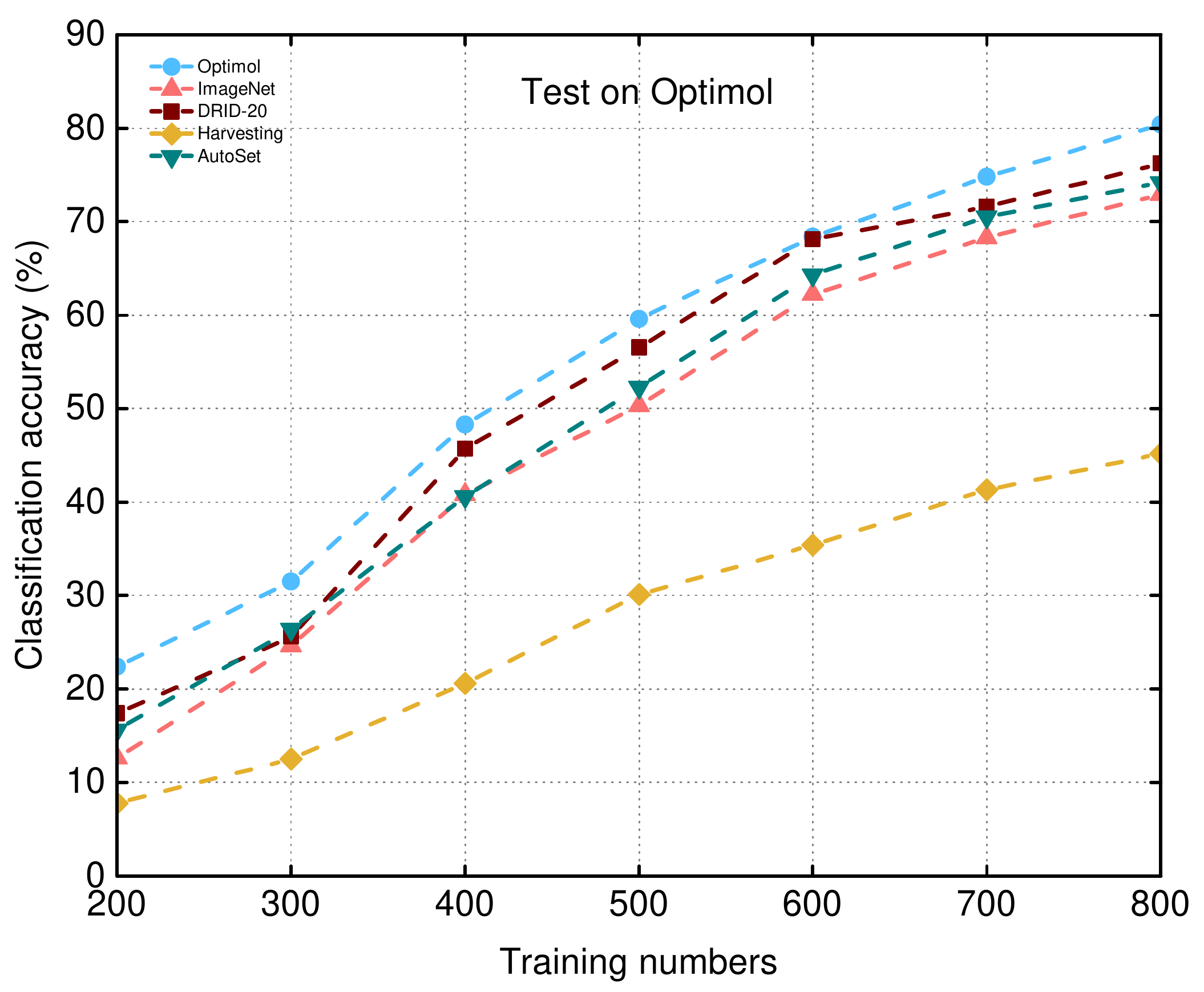}}
	\hspace{0.8cm}
	\subfloat[]{%
		\includegraphics[width=1.7in, height=1.6in]{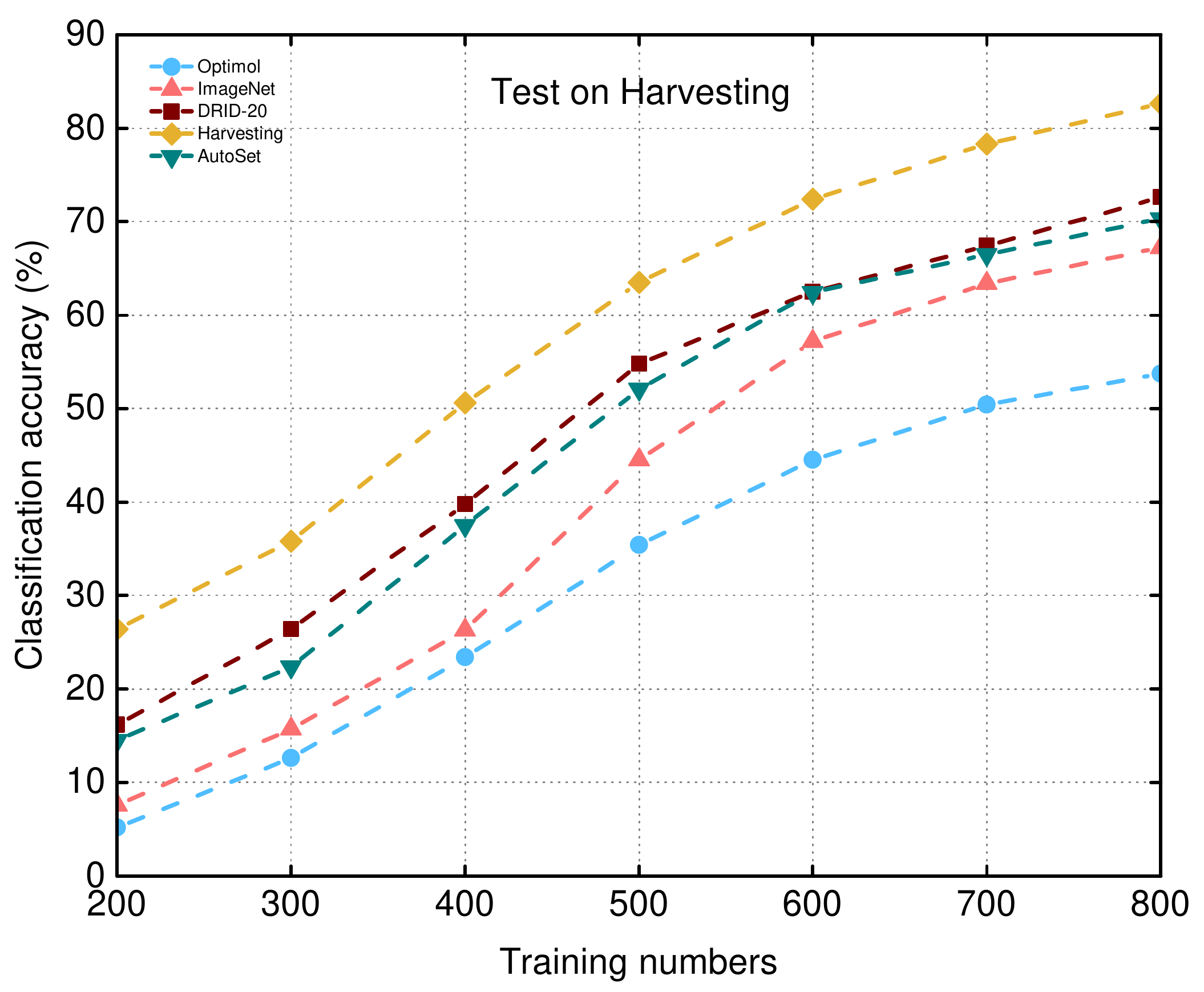}}
	\hspace{0.8cm}
	\subfloat[]{%
		\includegraphics[width=1.7in, height=1.6in]{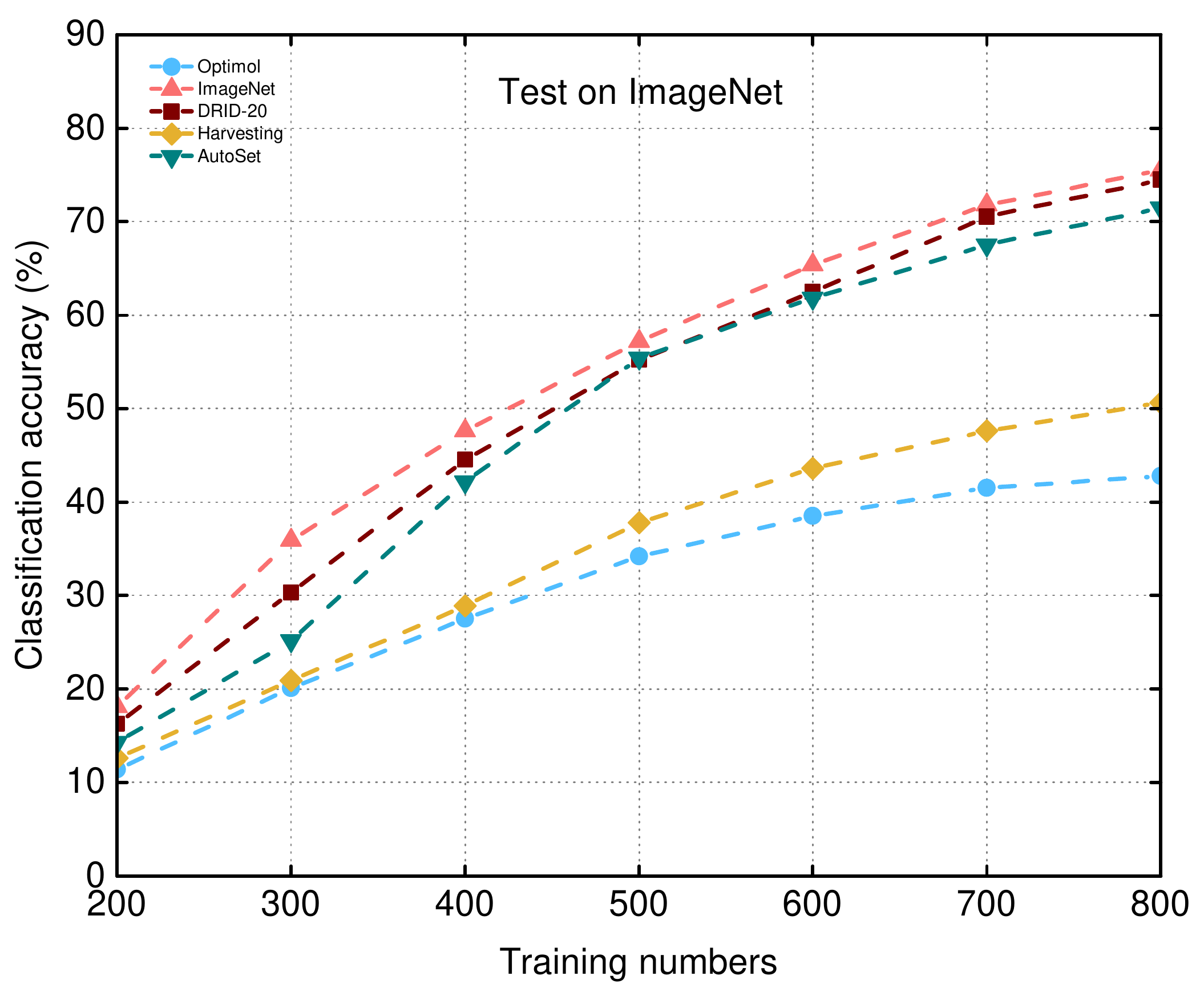}}\
	\subfloat[]{%
		\includegraphics[width=1.7in, height=1.6in]{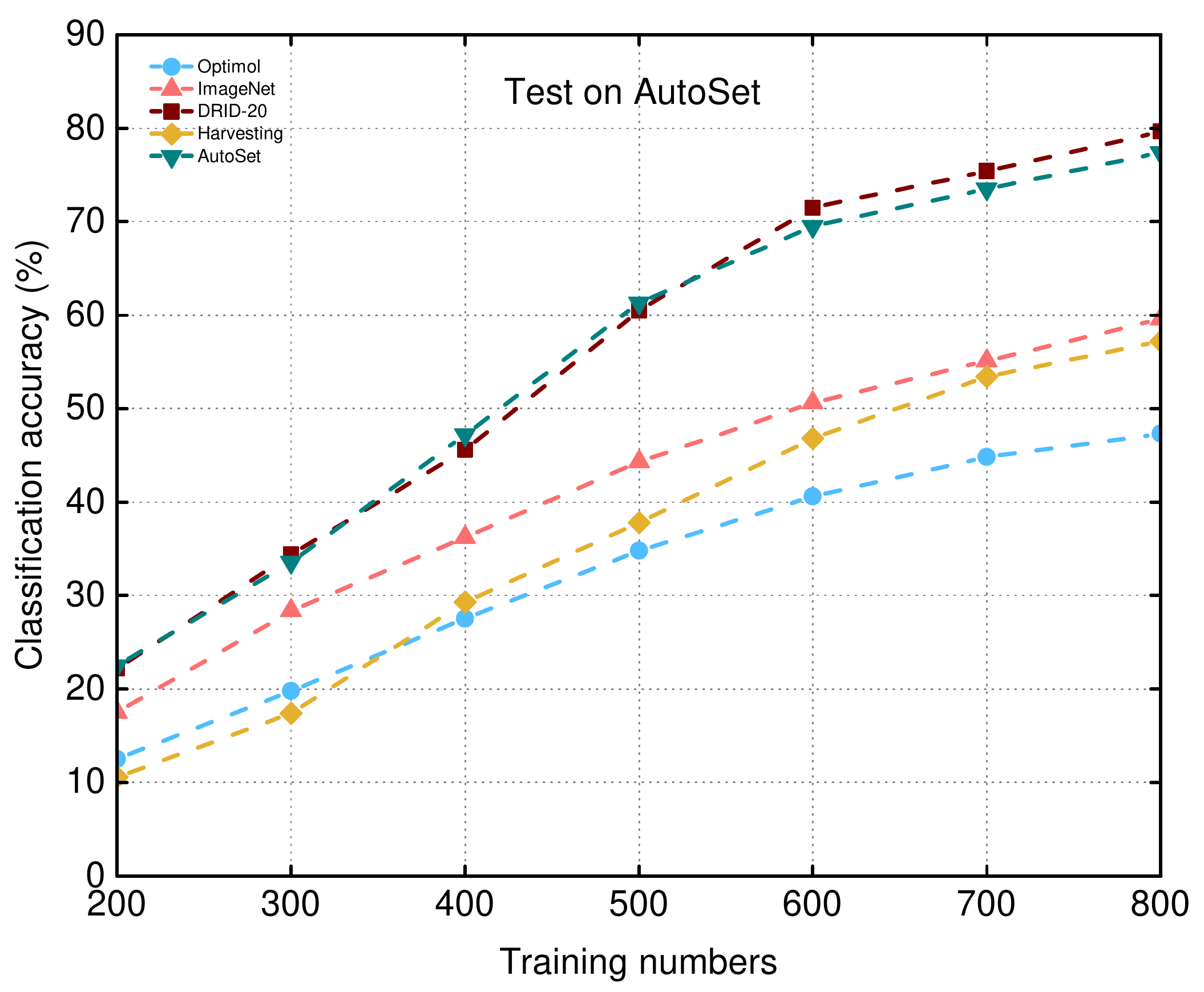}}
	\hspace{0.8cm}
	\subfloat[]{%
		\includegraphics[width=1.7in, height=1.6in]{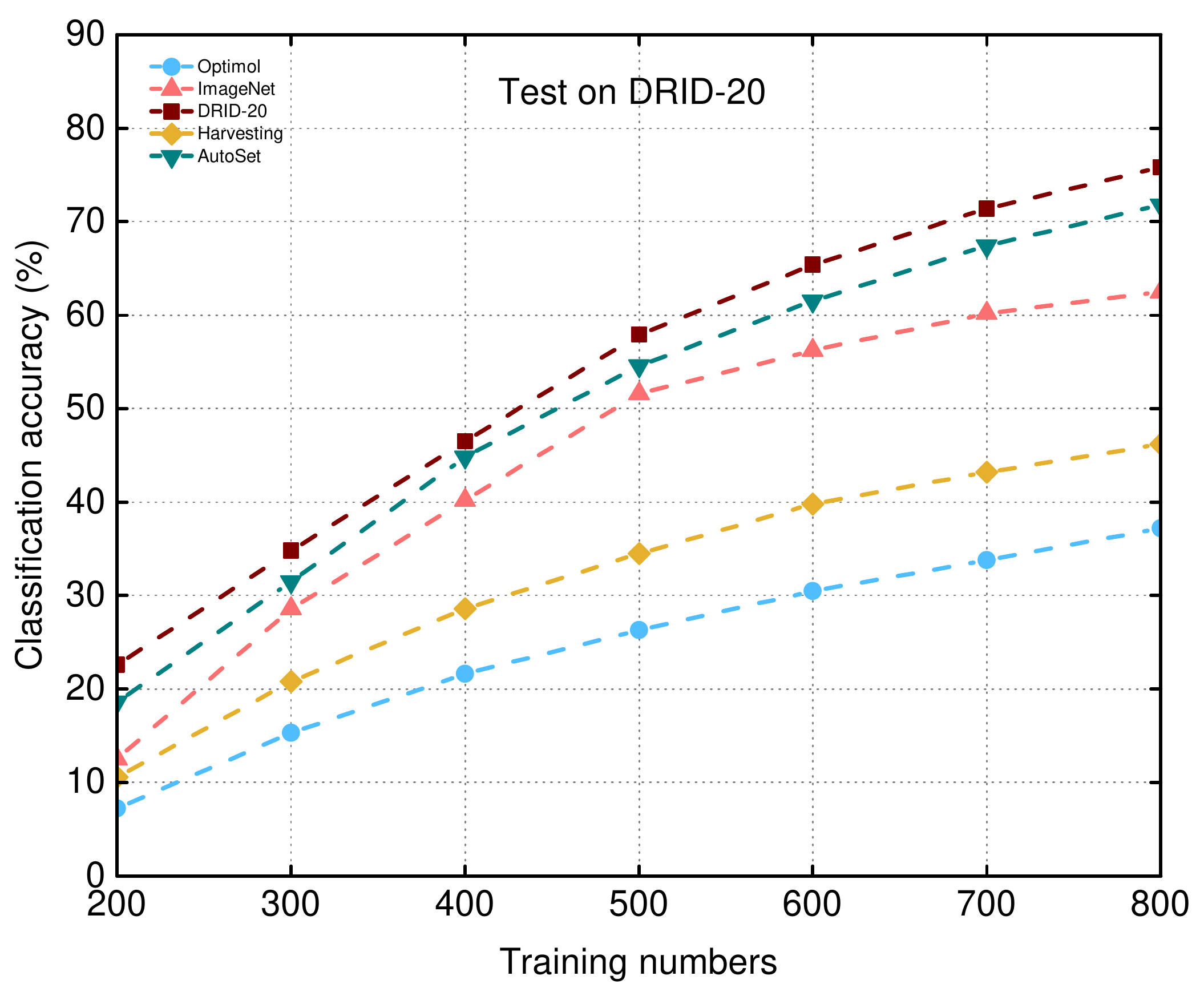}}
	\hspace{0.8cm}
	\subfloat[]{%
		\includegraphics[width=1.7in, height=1.6in]{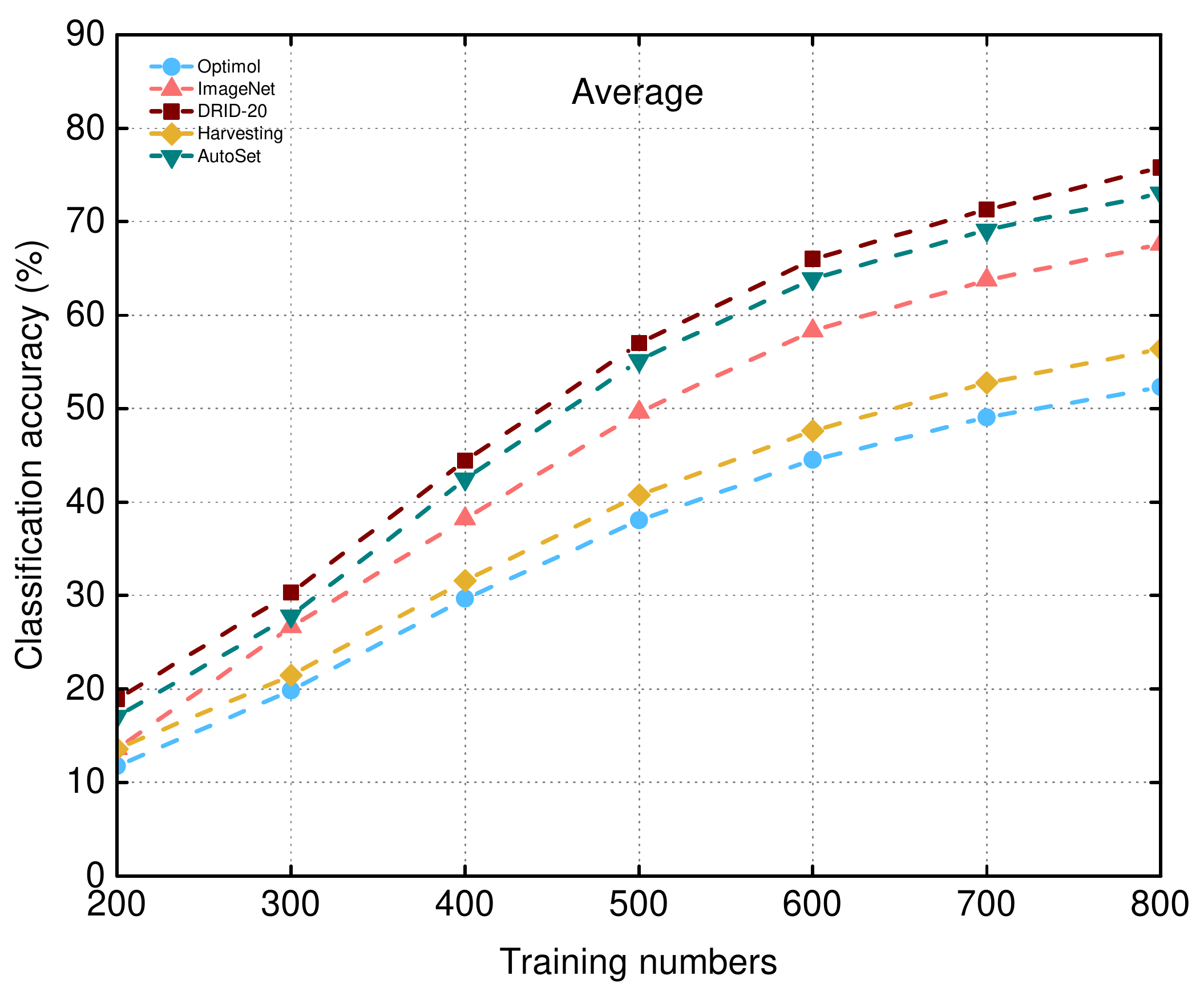}}
	\caption{Cross-dataset generalization ability of classifiers learned from Optimol, Harvesting, ImageNet, AutoSet, DRID-20 and then tested on: (a) Optimol, (b) Harvesting, (c) ImageNet, (d) AutoSet, (e) DRID-20, (f) Average.}
	\label{fig6}
\end{figure*}

For the comparison of cross-dataset generalization ability, we randomly select 200 images for each category as the testing data. For the choice of training data, we sequentially select [200,300,400,500,600,700,800] images per category from various datasets as the positive training samples, and use 1000 fixed irrelevant images as the negative training samples to learn the image classifiers. The training images in each category are selected randomly. In addition, the training data and testing data have no duplicates. Like the comparison of image classification ability, we also compare the category ``airplane", ``bird", ``cat", ``dog", ``horse" and ``car/automobile" among STL-10 \cite{coates2011analysis}, CIFAR-10
\cite{krizhevsky2009learning} and DRID-20. For comparison with ImageNet \cite{deng2009imagenet}, Optimol \cite{li2010optimol}, Harvesting \cite{schroff2011harvesting} and AutoSet \cite{icme2016yao}, we also use all the 20 categories among these datasets. The average classification accuracy represents the cross-dataset generalization ability of one dataset on another dataset. The experimental results are shown in Fig. \ref{fig5} and Fig. \ref{fig6} respectively.

For the comparison of dataset diversity, we select five common categories ``airplane", ``bird",``cat",``dog" and ``horse" in STL-10, ImageNet and DRID-20 as testing examples. 
Following method \cite{deng2009imagenet} and \cite{collins2008towards}, we compute the average image of each category and measure the lossless JPG file size. In particular, we resize all images in STL-10, ImageNet, DRID-20 to 32$\times$32 images, and create average images for each category from 100 randomly sampled images. Fig. \ref{fig7} (a) presents the lossless JPG file sizes of five common categories in dataset DRID-20, ImageNet and STL-10. The example and average images for five categories in three datasets are shown in Fig. \ref{fig7} (b).

For image classification ability and cross-dataset generalization ability comparison, we set the same options for all datasets. Particularly, we set the type of SVM as C-SVC, the type of kernel as a radial basis function and all other options as the default LIBSVM options. For all datasets, we extract the same dense histogram of oriented gradients (HOG) feature \cite{dalal2005histograms} and train one-versus-all classifiers. 

\subsubsection{Baselines}

In order to validate the performance of our dataset, we compare the image classification ability, cross-dataset generalization ability and dataset diversity of our dataset DRID-20 with two sets of baselines: 

$\bullet$ Manually labelled datasets. The manually labelled datasets include STL-10 \cite{coates2011analysis}, CIFAR-10 \cite{krizhevsky2009learning} and ImageNet \cite{deng2009imagenet}. The STL-10 dataset has ten categories, and each category of which contains 500 training images and 800 test images. All of the images are color 96 $\times$ 96 pixels. The CIFAR-10 dataset consists of 32$\times$32 images in 10 categories, with 6000 images per category. ImageNet is an image dataset organized according to the WordNet hierarchy. It provides an average of 1000 images to illustrate each category.

$\bullet$ Automated datasets. The automated datasets contain 
Optimol \cite{li2010optimol}, Harvesting \cite{schroff2011harvesting} and AutoSet \cite{icme2016yao}. For \cite{li2010optimol}, 1000 images for each category are collected by using the incremental learning method. Following \cite{schroff2011harvesting}, we firstly obtain the candidate images from the web search and rank the returned images by the text information. Then we use the top-ranked images to learn visual classifiers to re-rank the images once again. 
We select the categories in DRID-20 as the target queries and accordingly obtain the multiple textual metadata. Following the proposed method in \cite{icme2016yao}, we take iterative mechanisms for noisy images filtering and construct the dataset. In total, we construct 20 same categories as DRID-20 for Optimol, Harvesting and AutoSet.   
 
\subsubsection{Experimental results for image classification}

By observing Fig. \ref{fig3} and Fig. \ref{fig4}, we make the following conclusions:

It is interesting to observe that the categories ``airplane", ``tv" and ``plant"  have a relatively higher classification accuracy than other categories with a small amount of training data. A possible explanation is that the scenes and visual patterns of ``airplane", ``tv" and ``plant" are relatively simpler than other categories. Even with a small amount of training data, there is still a large number of positive patterns in both auxiliary and target domains. That is to say, the samples are densely distributed in the feature space, and the distribution of the 
two domains overlaps much more easily. 

\begin{figure*} [t]
	\centering
	\subfloat[]{\includegraphics[width=1.6in, height=1.6in]{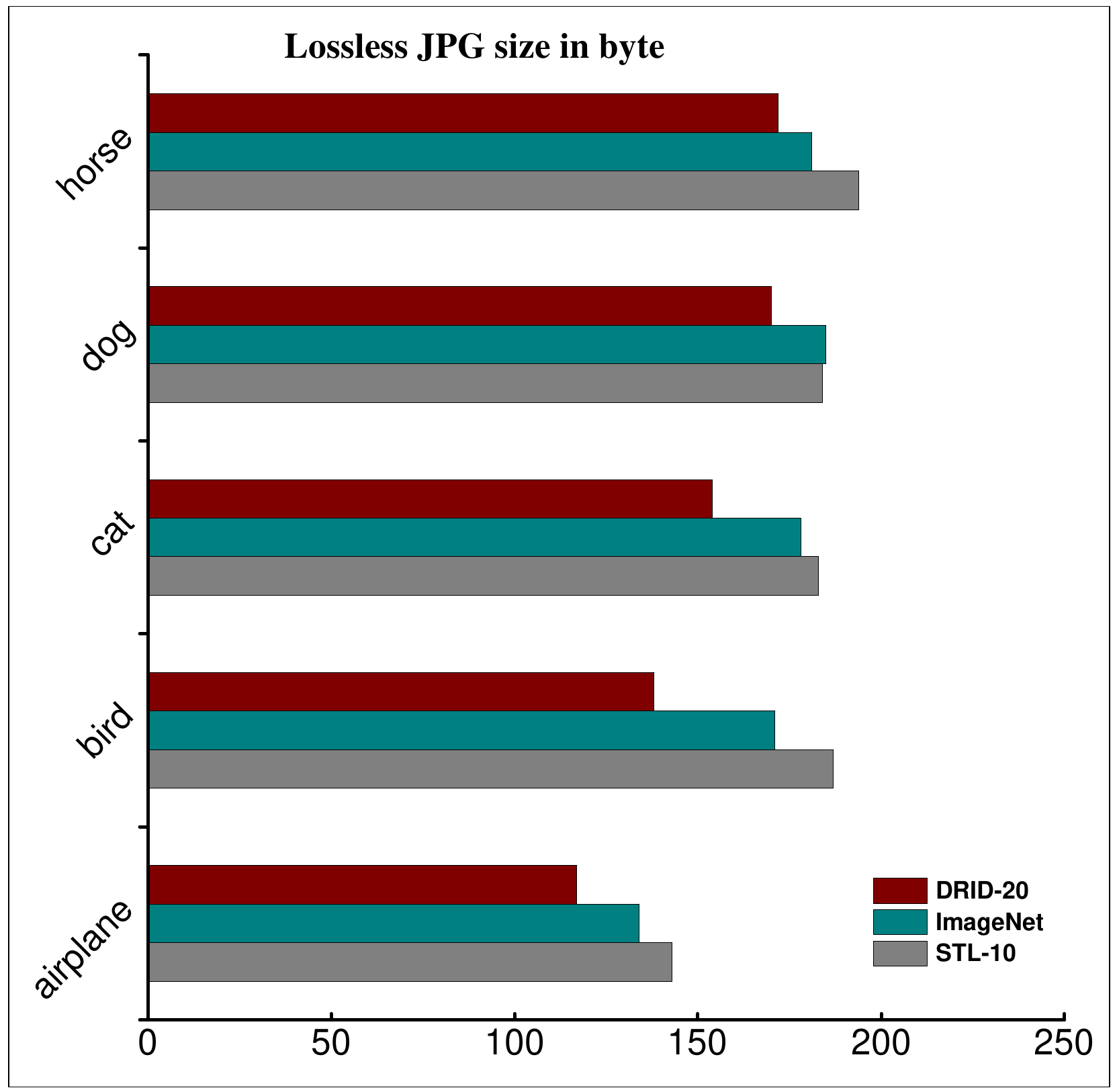}}
	\subfloat[]{\includegraphics[width=5.4in, height=1.6in]{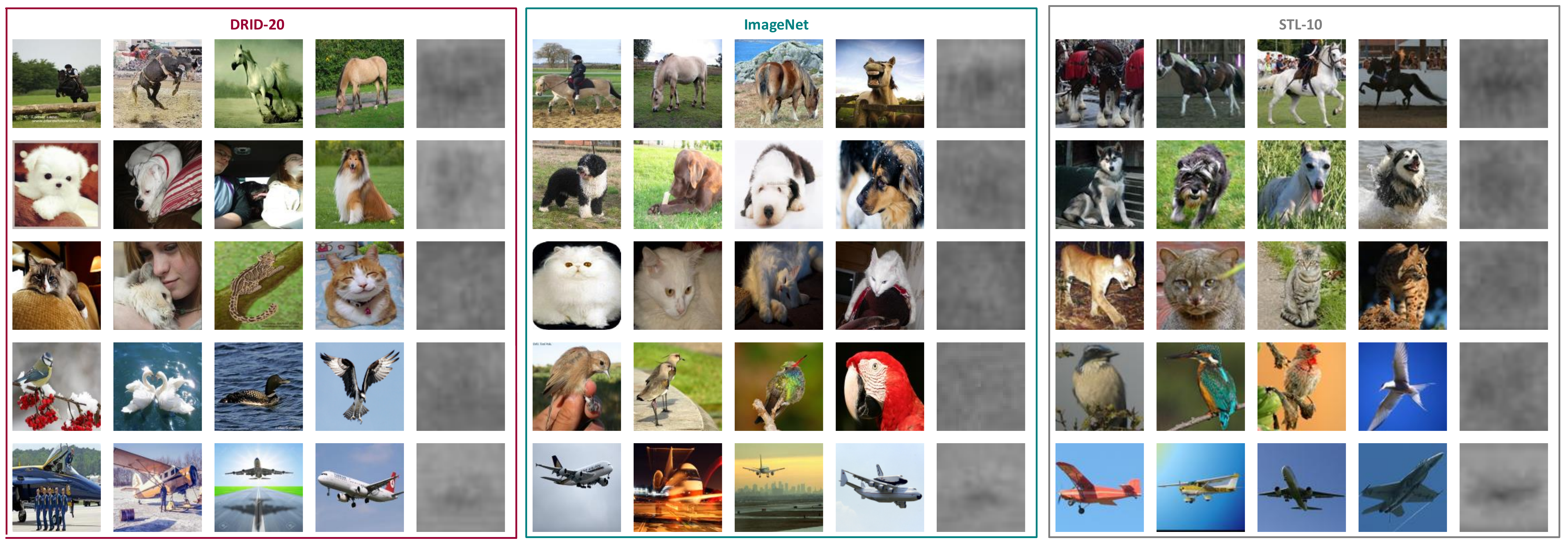}}
	\caption{(a) Comparison of the lossless JPG file sizes of average images for five different categories in DRID-20, ImageNet and STL-10. (b) Example images from DRID-20, ImageNet, STL-10 and average images for each category indicated by (a).}
	\label{fig7}
\end{figure*}

CIFAR-10 exhibits a much worse performance on image classification than STL-10 and DRID-20 according to its accuracy over six common categories. This demonstrates that the classifier learned with the training data from the auxiliary domain performs poorly on the target domain. The explanation is perhaps that the data distributions of CIFAR-10 are quite 
different from those of the PASCAL VOC 2007 dataset. The CIFAR-10 dataset has a more serious dataset bias problem than STL-10 and DRID-20.

STL-10 performs much better on category ``dog" than CIFAR-10 and DRID-20 when the number of training data is 400. The explanation is that STL-10 may have more effective visual patterns than CIFAR-10 and DRID-20 on category ``dog" with 400 training data. On the other hand, the positive samples from CIFAR-10 and DRID-20 are distributed sparsely in the feature space with 400 training images. It is likely that there are less overlap between the auxiliary and target domains for CIFAR-10 and DRID-20.

DRID-20 outperforms the automated datasets in terms of average accuracy in 20 categories, which demonstrates the domain robustness of DRID-20. A possible explanation is that our DRID-20 dataset, being constructed by multiple query expansions, has many more visual patterns or feature distributions than Harvesting and Optimol. At the same time, compared to AutoSet which uses iterative mechanisms in the process of image selection, MIL mechanisms can maximize the retention of useful visual patterns. Thus, our dataset has a better image classification ability.

\subsubsection{Experimental results for cross-dataset generalization}

Cross-dataset generalization measures the performance of classifiers learned from one dataset and tested on another dataset. It indicates the robustness of dataset \cite{torralba2011unbiased,mm2016yao}.
By observing Fig. \ref{fig5} and Fig. \ref{fig6}, we draw the following conclusions:

Compared to STL-10 and DRID-20, CIFAR-10 has a poor cross-dataset generalization ability except on its own dataset. The explanation is that the data distributions of its auxiliary domain and target domain are strongly related, making it difficult for other datasets to exceed its performance when tested on CIFAR-10. All images in CIFAR-10 are cut to 32$\times$32 and the objects in these images are located in the middle of the image. Besides, these images contain relatively fewer other objects and scenes. The images in STL-10 are 96$\times$96 and are full size in DRID-20. These images not only contain target objects, but also include a large number of other scenarios and objects. Based on these conditions, CIFAR-10 has a serious dataset bias problem which coincides with its average cross-dataset generalization performance.

AutoSet is better than Optimol, Harvesting and ImageNet but slightly worse than DRID-20, possibly because the distribution of samples is relatively rich. AutoSet is constructed using multiple textual meta-data and the objects of its images have variable appearances, positions, viewpoints, and poses. 

DRID-20 outperforms CIFAR-10, STL-10, ImageNet, Optimol, Harvesting and AutoSet in terms of average cross-dataset performance, which demonstrates the domain robustness of DRID-20. This may be because DRID-20 constructed by multiple query expansions and MIL selection mechanisms has much more effective visual patterns than other datasets given the same number of training samples. In other words, DRID-20 has a much richer feature distribution and is more easily overlapped with unknown target domains.

\subsubsection{Experimental results for dataset diversity}

The lossless JPG file size of the average image for each category reflects the amount of information in an image. The basic idea is that a diverse image dataset will result in a blurrier average image, the extreme being a gray image. Meanwhile, an image dataset with limited diversity will result in a more structured, sharper average image. Therefore, we expect the average image of a more diverse image dataset to have a smaller JPG file size. By observing Fig. \ref{fig7}:

DRID-20 has a slightly smaller JPG file size than ImageNet and STL-10 which indicates the diversity of our dataset. This phenomenon is universal for all five categories. It can be seen that the average image of DRID-20 is blurred and it is difficult to recognize the object, while the average image of ImageNet and STL-10 is relatively more structured and sharper.

DRID-20 is constructed with the goal that images in this dataset should exhibit domain robustness and be able to effectively alleviate the dataset bias problem. To achieve domain robustness, we not only consider the source of the candidate images, but also retain the images from different distributions.

\subsection{Comparison of Object Detection Ability}

The idea of training detection models without bounding boxes has received renewed attention due to the success of the DPM \cite{felzenszwalb2010object} detector. To compare the object detection ability of our collected data with other baseline methods \cite{divvala2014learning,felzenszwalb2010object,siva2011weakly,prest2012learning,icme2016yao}, we selected PASCAL VOC 2007 as the test data. The reason is recent state-of-the-art weakly supervised and web-supervised methods have been evaluated on this dataset.
 
\subsubsection{Experimental setting} 

For each query expansion, we train a separate DPM to constrain the visual variance. We resize images to a maximum of 500 pixels and ignore images with extreme aspect ratios (aspect ratio $>$ 2.5 or $<$ 0.4). To avoid getting stuck to the image boundary during the latent re-clustering step, we initialize our bounding box to a sub-image within the image that ignores the image boundaries. Following \cite{felzenszwalb2010object}, we also initialize components using the aspect-ratio heuristic. Some of the components across different query expansion detectors ultimately learn the same visual pattern. For example, the images corresponding to the query expansion ``walking horse" are similar to the images corresponding to ``standing horse". In order to select a representative subset of the components and merge similar components, we represent the space of all query expansions components by a graph $G=\left \{ C,E \right \}$, in which each node represents a component and each edge represents the visual similarity between them. 
The score $d_{i}$ for each node corresponds to the average precision. The weight on each edge $e_{i,j}$ is obtained by running the $j\mathrm{th}$ component detector on the $i\mathrm{th}$ component set. We solve the same objective function proposed in \cite{divvala2014learning} to select the representative components $ S $ ($S\subseteq V$) :
\begin{equation}\label{eq22}
\max_{S}\sum_{i\in V}d_{i}\cdot \vartheta (i,S)  \\
\end{equation}
where $\vartheta$ is a soft coverage function that implicitly pushes for diversity:
\begin{equation}\label{eq23}
\vartheta(i,S)=
\begin{cases}
1 &\mbox{$i\in S$}\\
1-\prod _{j\in S}(1-e_{i,j}) &\mbox{$ i \notin S $}.
\end{cases}
\end{equation}
After the representative subset of components has been obtained, we augment them with the method described in \cite{felzenszwalb2010object} and subsequently merge all the components to produce the final detector.

\subsubsection{Baselines}

In order to validate the object detection ability of our collected data, we compare our approach with three sets of baselines:

$\bullet$ Weakly supervised methods. The weakly supervised learning methods include WSL \cite{siva2011weakly} and SPM-VID  \cite{prest2012learning}. WSL uses weak human supervision (VOC data with image-level labels for training) and initialization from objectness. SPM-VID is trained on manually selected videos without bounding boxes and shows results in 10 out of 20 categories.  

$\bullet$ Web-supervised methods. Such methods include WSVCL \cite{divvala2014learning} and IDC-MTM \cite{icme2016yao}. WSVCL takes web supervision and then trains a mixture DPM detector for the object. IDC-MTM collects candidate images with multiple textual metadata and filters these images using an iterative method. Images which are not filtered out are then selected as positive training images for mixture DPM detector learning.       

$\bullet$ Fully supervised method. The fully supervised method includes OD-DPM \cite{felzenszwalb2010object}. OD-DPM is a fully supervised object detection method and it is a possible upper bound for weakly supervised and web-supervised approaches. 

\subsubsection{Experimental results for object detection}

We report the performance of object detection on PASCAL VOC 2007 test set. Table \ref{tab2} shows the results of our proposed method and 
\begin{table}[htb]
	\centering
	\caption{OBJECT DETECTION RESULTS (A.P.) (\%) ON PASCAL VOC 2007 (TEST).}
	\renewcommand{\arraystretch}{1.2}
	\begin{tabular}{|c|c|c|c|c|c|c|c|}
		\hline
		Method & \cite{siva2011weakly} & \cite{prest2012learning} & \cite{divvala2014learning} & \cite{icme2016yao} & $\mathbf{Our}$ & \cite{felzenszwalb2010object} \\
		\hline
		Supervision & weak & weak & web & web & web & full\\
		\hline
		\hline		 
		airplane & 13.4 & $\mathbf{17.4}$ & 14.0 & 14.8 & 15.5 & 33.2\\
		\hline	
		bike  & $\mathbf{44.0}$ & - & 36.2 & 38.4 & 40.6 & 59.0\\
		\hline 
		bird  & 3.1 & 9.3 & 12.5 & \textbf{16.5} & 16.1 & 10.3\\
		\hline	
		boat & 3.1 & 9.2 & $\mathbf{10.3}$ & 7.4 & 9.69 & 15.7 \\
		\hline
		bottle & 0.0 & - & 9.2 & 12.6 & $\mathbf{13.7}$ & 26.6\\
		\hline	
		bus & 31.2 & - & 35.0 & 39.5 & $\mathbf{42.0}$ & 52.0 \\
		\hline
		car & $\mathbf{43.9}$ & 35.7 & 35.9 & 38.1 & 37.9 & 53.7\\
		\hline	
		cat & 7.1 & 9.4 & 8.4 & 8.9 & $\mathbf{9.8}$ & 22.5 \\
		\hline
		chair & 0.1 & - & $\mathbf{10.0}$ & 9.3 & 9.6  & 20.2\\
		\hline	
		cow & 9.3 & 9.7 & 17.5 & 17.9 & $\mathbf{18.4}$ & 24.3 \\		
		\hline
		table & 9.9 & - & 6.5 & 10.2 & $\mathbf{10.6}$ & 26.9\\
		\hline	
		dog & 1.5 & 3.3 & $\mathbf{12.9}$ & 11.5 & 11.6& 12.6\\
		\hline 
		horse & 29.4 & 16.2 & 30.6 & 31.8 & $\mathbf{36.1}$& 56.5 \\
		\hline	
		motorcycle& $\mathbf{38.3}$ & 27.3 & 27.5 & 29.7 & 36.9 & 48.5 \\
		\hline
		person & 4.6 & - & 6.0 & 7.2 & $\mathbf{7.9}$ & 43.3\\
		\hline	
		plant & 0.1 & - & $\mathbf{1.5}$ & 1.1 & 1.3 & 13.4\\
		\hline
		sheep & 0.4 & - & 18.8 & 19.5 & $\mathbf{20.4}$ & 20.9 \\
		\hline	
		sofa & 3.8 & - & 10.3 & 10.3 & $\mathbf{10.8}$ & 35.9 \\
		\hline
		train & $\mathbf{34.2}$ & 15.0 & 23.5 & 24.2 & 27.6 & 45.2 \\
		\hline	
		tv/monitor & 0.0 & - & 16.4 & 15.6 & $\mathbf{18.4}$  & 42.1\\				
		\hline
		average & 13.87 & 15.25 & 17.15 & 18.22 & $\mathbf{19.74}$  & 33.14 \\				
		\hline 	    	    	    	    	    	     	       		
	\end{tabular}
	\label{tab2}
\end{table}
compares it to the state-of-the-art weakly supervised and web-supervised methods \cite{siva2011weakly,prest2012learning,divvala2014learning,icme2016yao}. By observing the Table \ref{tab2}, we draw the following conclusions:  

Compared to WSL and SPM-VID (which use weak supervision) and OD-DPM (which uses full supervision), the training sets of our proposed approach and WSVCL, IDC-MTM do not need to be labelled manually. Nonetheless, the results of our proposed approach and WSVCL, IDC-MTM surpass the previous best results of weakly supervised object detection methods WSL, SPM-VID. A possible explanation is perhaps that both our approach and that of WSVCL, IDC-MTM use multiple query expansions for candidate image collection, and the training data collected by our approach and WSVCL, IDC-MTM are richer and contain more effective visual patterns. 

In most cases, our method surpasses the results obtained from WSVCL, IDC-MTM, which also uses web supervision and multiple query expansions for candidate images collection. The explanation for this is that we use different mechanisms for the removal of noisy images. Compared to WSVCL, IDC-MTM which uses iterative mechanisms in the process of noisy images filtering, our approach applies an MIL method for removing noisy images. This maximizes the ability to retain images from different data distributions while filtering out the noisy images. 

Our approach outperforms the weakly supervised and web-supervised methods \cite{siva2011weakly,prest2012learning,divvala2014learning,icme2016yao}. The main reason being that our training data is generated using multiple expansions and MIL filtering mechanisms. Thus, our data contains much richer and more accurate visual descriptions for these categories. In other words, our approach discovers much more useful linkages to visual patterns for the given category.

\begin{figure} [t]
	\centering
	\subfloat[]{%
		\includegraphics[width=1.7in, height=1.6in]{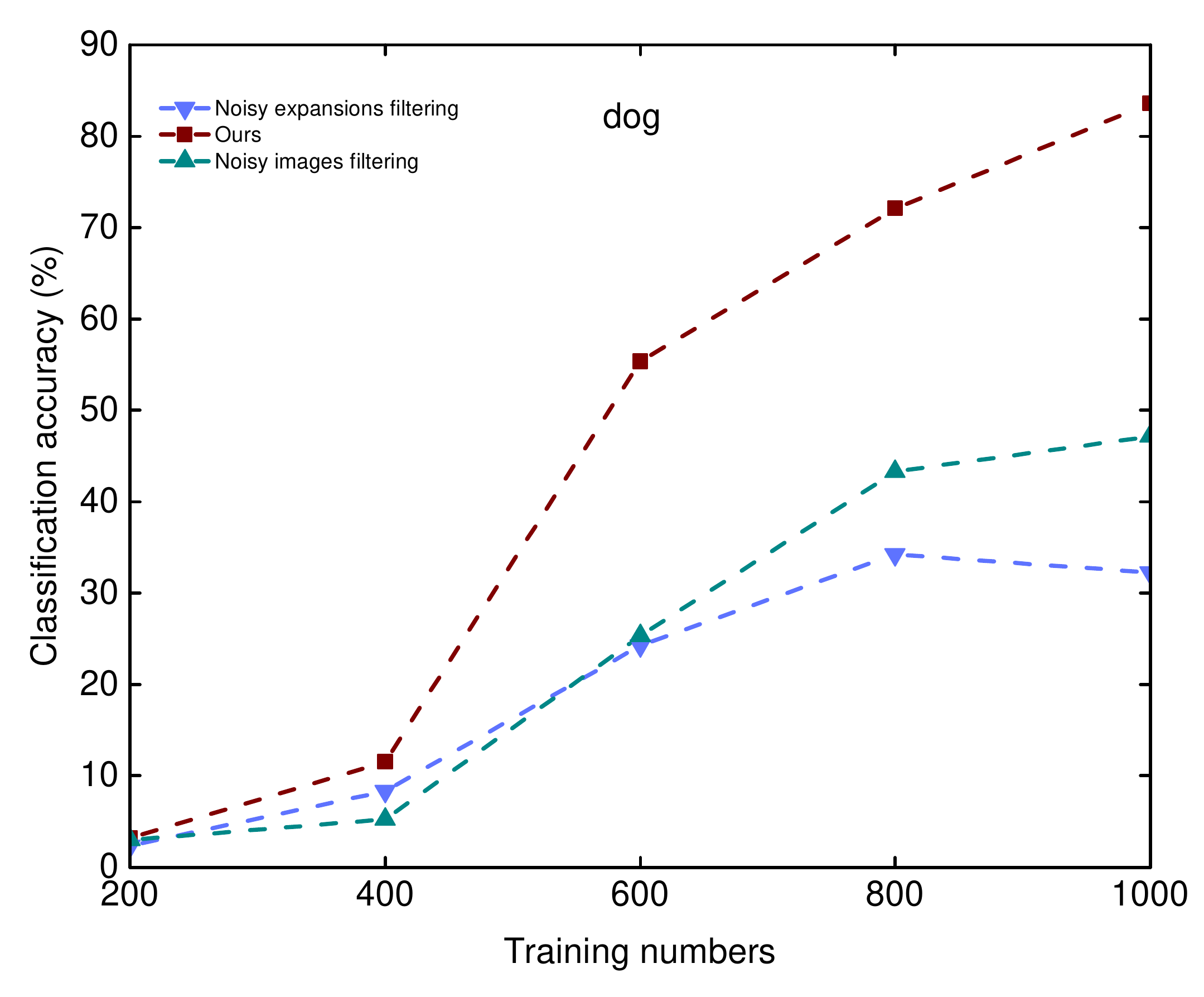}}
	\subfloat[]{%
		\includegraphics[width=1.7in, height=1.6in]{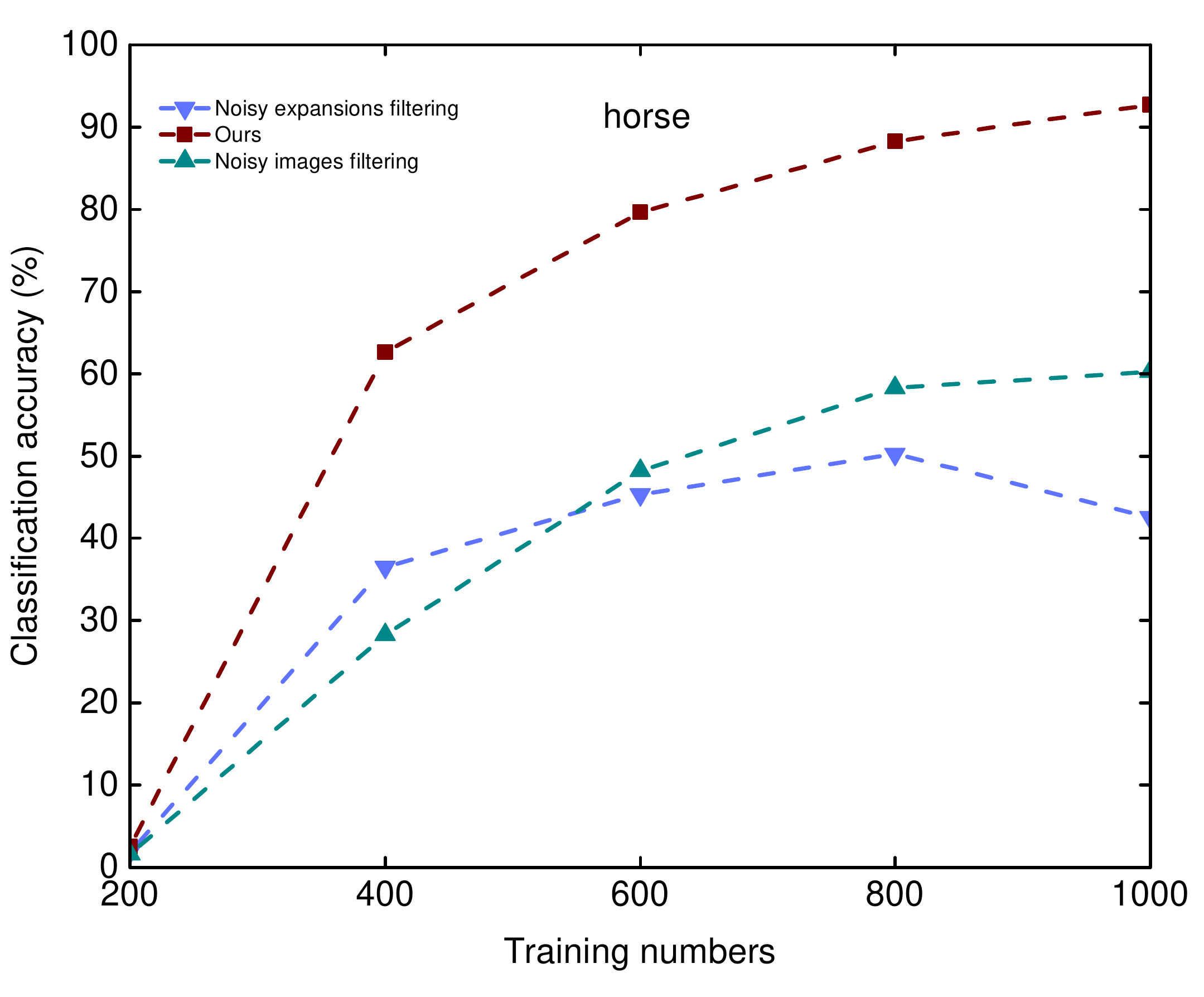}}
	\caption{The image classification ability of different steps on category (a) dog, (b) horse.}
	\label{fig8}
\end{figure}

\subsection{Different Steps Analysis}

Our proposed framework involves three major steps: query expanding, noisy expansions filtering and noisy images filtering. In order to quantify the role of different steps contributing to the final results, we build two new datasets. One is based on noisy expansions filtering and another one is based on noisy images filtering. In particular, we construct the noisy expansions filtering based image dataset by query expanding and noisy expansions filtering. After the noisy expansions are filtered out, we retrieve the top images from image search engine for selected expansions to construct the dataset. We build the noisy images filtering based image dataset by query expanding and noisy images filtering. After we obtain the query expansions through query expanding, we take MIL based methods for noisy images filtering and construct the dataset. 

We compare the image classification ability of these two new datasets with our DRID-20. By selecting ``dog" and ``horse" as two target categories to construct the dataset, we sequentially collect [200,400,600,800,1000] images for each category and test the image classification ability on PASCAL VOC 2007 dataset. The results are shown in Fig. \ref{fig8}. It can be seen that:

When the number of images in each category is below 600, the noisy expansions filtering based method tends to have a better image classification ability. This is possibly because the top images returned from the image search engine have a relatively high accuracy. The noisy images induced by noisy expansions are more serious than those caused by the image search engine. With the increased number of images in each category, the images returned from the image search engine have more and more noisy images. The noisy images caused by image search engine have a worse effect than those induced by noisy expansions.

Our method outperforms both noisy expansions filtering based and noisy images filtering based method. This is because our method, which takes a combination of noisy expansions filtering and noisy images filtering, can effectively remove the noisy images induced by both noisy expansions and the image search engine. 

\begin{figure} [t]
	\centering
	\subfloat[]{%
		\includegraphics[width=1.7in, height=1.6in]{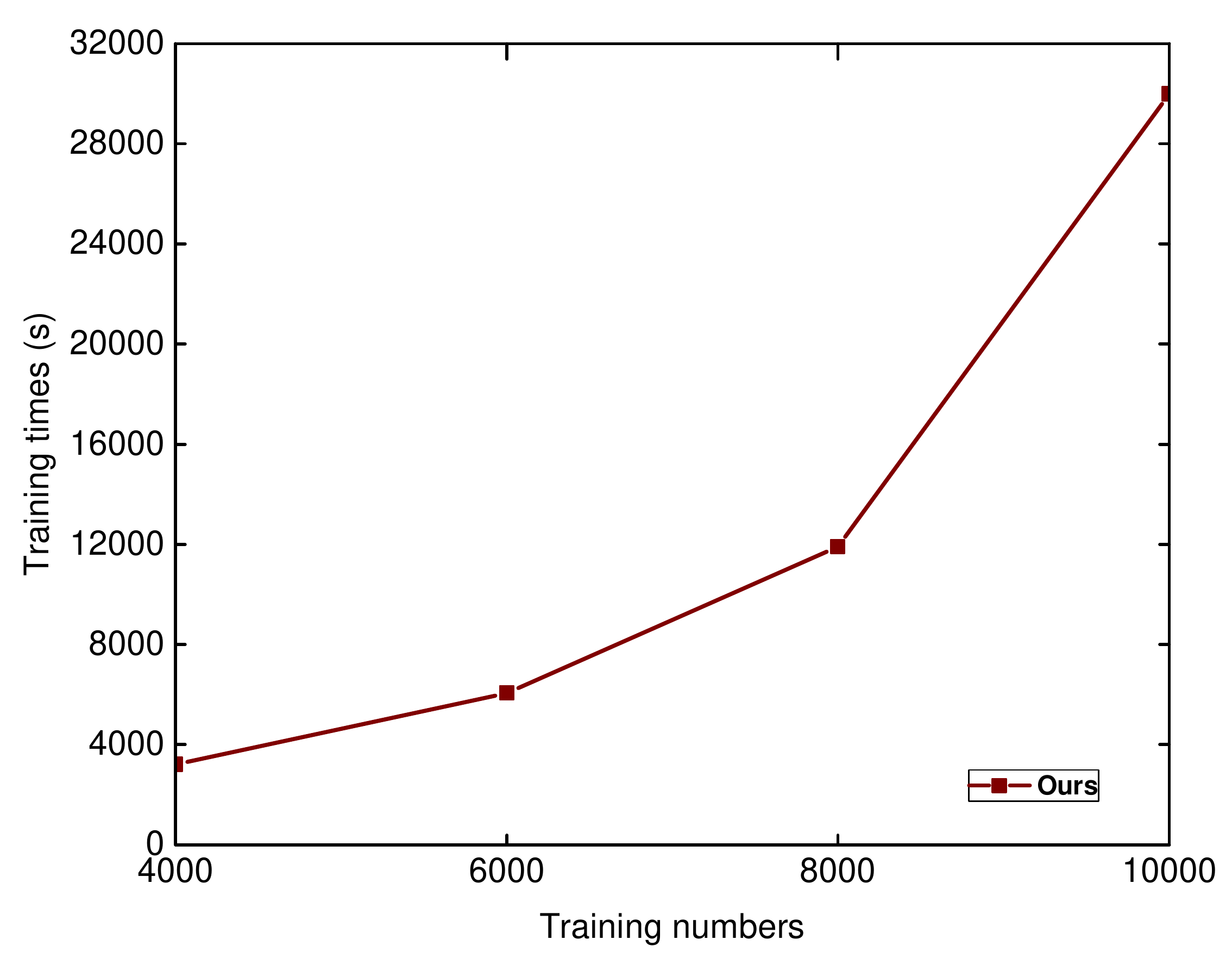}}
	\subfloat[]{%
		\includegraphics[width=1.7in, height=1.6in]{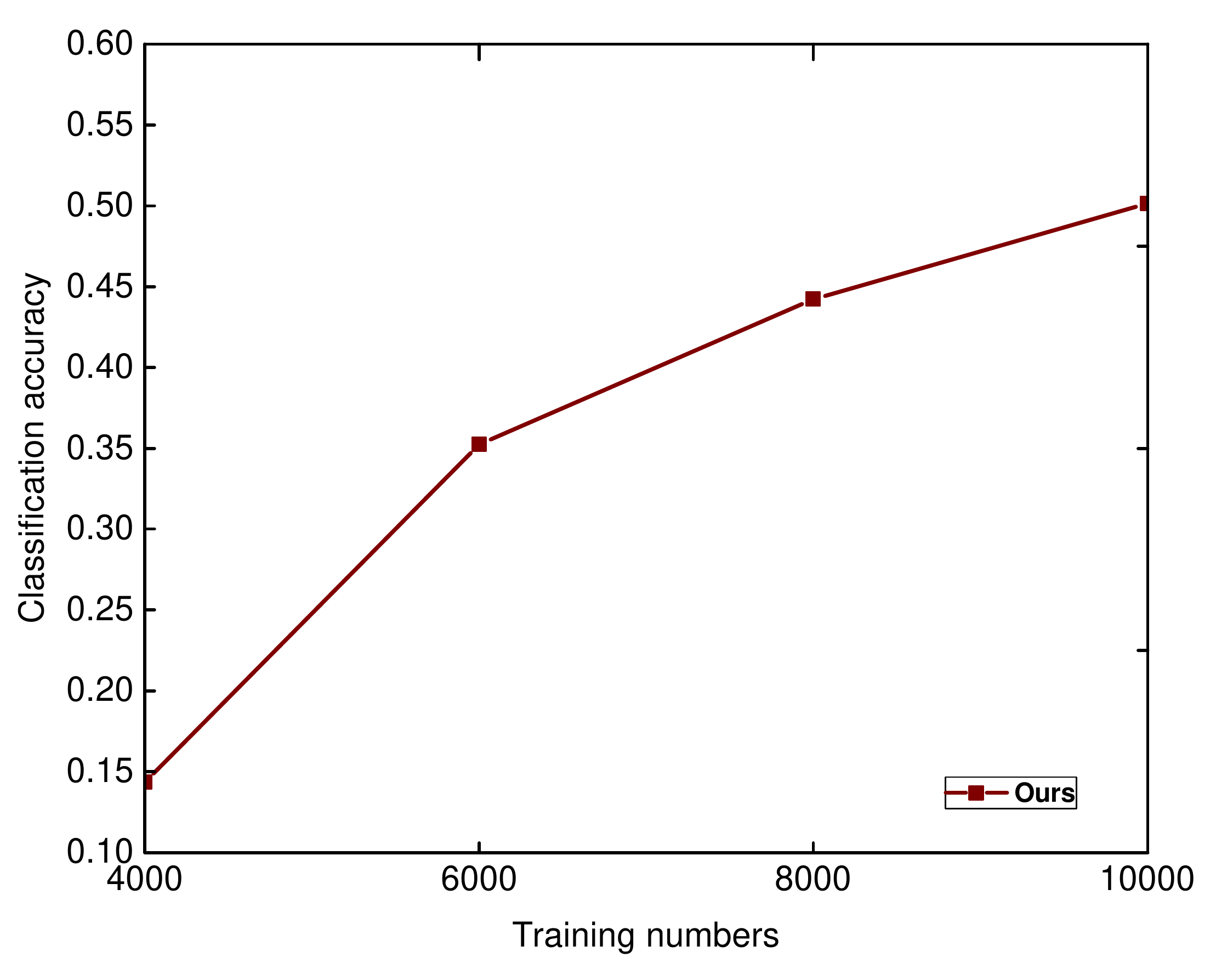}}
	\caption{The training time and image classification accuracies with respect to various numbers of training images.}
	\label{fig9}
\end{figure}

\subsection{Time Complexity And Space Complexity Analysis}

For the time complexity analysis, we mainly focus on multi-instance learning time. During the process of MIL learning, we solve the convex problem in \eqref{eq10} by using the cutting-plane algorithm. By identifying the most violating candidate and solving the MKL sub-problem at each iteration, the time complexity of \eqref{eq10} can be approximately computed as $ T\cdot O $(MKL), where \textit{T} is the number of iterations and $ O $(MKL) is the time complexity of the MKL sub-problem. According to [29], the time complexity of MKL is between $ t \cdot O(LCM) $ and $ t \cdot O((LCM)^{2.3}) $, where $ M, L, C  $ are the numbers of latent domains, bags and categories respectively. $ t $ is the number of iterations in MKL.
We have chosen PASCAL VOC 2007 as the testing set for the evaluation of our method. In particular, we use various numbers of training images for each category to learn the classifier. PASCAL VOC 2007 has 20 categories and we use $ n $ training images for each category, so we have a total of 20$ n $ training images. Fig. \ref{fig9} shows the training time and image classification accuracies with respect to the various numbers of training images. We can observe that both training time and image classification accuracies increase as the number of training images grows.

\section{Conclusion and future work}

In this paper, we presented a new framework for domain-robust image dataset construction with web images. Three successive modules were employed in the framework, namely query expanding, noisy expansion filtering and noisy image filtering. To verify the effectiveness of our proposed method, we constructed an image dataset DRID-20. Extensive experiments have shown that our dataset not only has better domain adaptation ability than the traditional manually labelled datasets STL-10, CIFAR-10 and ImageNet, but also has better domain adaptation ability than the automated datasets Optimol, Harvesting and AutoSet. In addition, our data was successfully applied to help improve object detection on PASCAL VOC 2007, and the results demonstrated the superiority of our method to several weakly supervised and web-supervised state-of-the-art methods. We have publicly released the DRID-20 dataset to facilitate the research in this field. 

\ifCLASSOPTIONcaptionsoff
  \newpage
\fi

\end{document}